\pdfoutput=1

\documentclass{article}
\usepackage{microtype}
\usepackage{graphicx}
\usepackage{booktabs}
\usepackage{tabularx}
\usepackage{graphicx} 
\usepackage{float}
\usepackage{wrapfig}
\newcommand{\Eqref}[1]{Eq.~(\ref{#1})}
\usepackage{amssymb}
\usepackage{multirow}
\usepackage{hyperref}

\PassOptionsToPackage{numbers,compress}{natbib}
\usepackage[final]{Formatting_Instructions_For_NeurIPS_2025/neurips_2025}
\usepackage{comment}

\usepackage[utf8]{inputenc}
\usepackage[T1]{fontenc}

\usepackage{amsmath}
\usepackage{amssymb}
\usepackage{mathtools}
\usepackage{caption}
\usepackage{amsthm}
\usepackage{adjustbox}
\usepackage[capitalize,noabbrev]{cleveref}
\usepackage[textsize=tiny]{todonotes}

\usepackage{xcolor}
\usepackage{hyperref}
\usepackage{tikz}
\usepackage{amsmath,amsfonts,bm}

\def\1{\bm{1}}

\DeclareMathAlphabet{\mathsfit}{\encodingdefault}{\sfdefault}{m}{sl}
\SetMathAlphabet{\mathsfit}{bold}{\encodingdefault}{\sfdefault}{bx}{n}

\theoremstyle{plain}

\theoremstyle{definition}

\theoremstyle{remark}

\definecolor{mydarkblue}{rgb}{0,0.08,0.45}
\hypersetup{
    colorlinks=true,
    citecolor= blue!70,
    linkcolor= mydarkblue,
    urlcolor= mydarkblue
}
\usetikzlibrary{trees} %
\usepackage[edges]{forest}
\tikzset{%
    parent/.style =          {align=center,text width=2cm,rounded corners=3pt, line width=0.3mm},%
    child/.style =           {align=center,text width=2.3cm,rounded corners=3pt, fill=blue!10,draw=blue!80,line width=0.3mm},
    grandchild/.style =      {align=center,text width=2cm,rounded corners=3pt},
    greatgrandchild/.style = {align=center,text width=1.5cm,rounded corners=3pt},
    greatgrandchild2/.style = {align=center,text width=1.5cm,rounded corners=3pt},    
    referenceblock/.style =  {align=center,text width=1.5cm,rounded corners=2pt},
    pretrain/.style =           {align=center,text width=2.9cm,rounded corners=3pt, fill=blue!10,draw=blue!80,line width=0.3mm},   
    pretrain_work/.style =           {align=center, text width=2.9cm,rounded corners=3pt, fill=blue!10,draw=blue!0,line width=0.3mm},  
    template/.style =           {align=center,text width=2.9cm,rounded corners=3pt, fill=red!10,draw=red!80,line width=0.3mm},   
    template_work/.style =           {align=center,text width=2.9cm,rounded corners=3pt, fill=red!10,draw=red!0,line width=0.3mm},    
    answer/.style =           {align=center,text width=2.9cm,rounded corners=3pt,line width=0.3mm},%
    answer_work/.style =           {align=center,text width=2.9cm,rounded corners=3pt,line width=0.3mm},%
    multiple/.style =           {align=center,text width=2.9cm,rounded corners=3pt, fill= orange!10,draw= orange!80,line width=0.3mm},   
    multiple_work/.style =           {align=center,text width=2.9cm,rounded corners=3pt, fill= orange!10,draw= orange!0,line width=0.3mm},        
    tuning/.style =           {align=center,text width=2.9cm,rounded corners=3pt, fill= magenta!10,draw= magenta!80,line width=0.3mm},   
    tuning_work/.style =           {align=center,text width=2.9cm,rounded corners=3pt, fill= magenta!10,draw= magenta!0,line width=0.3mm},
    assurance/.style =           {align=center,text width=2.9cm,rounded corners=3pt, fill= green!10,draw= green!80,line width=0.3mm},   
    assurance/.style =           {align=center,text width=2.9cm,rounded corners=3pt, fill= green!10,draw= green!0,line width=0.3mm},
}
\usepackage{color}
\usepackage{colortbl}
\cellcolor{gray!0}
\definecolor{Gray}{gray}{0.9}
\usepackage{xcolor}

\usepackage{etoc}
\etocdepthtag.toc{mtchapter}
\etocsettagdepth{mtchapter}{subsection}
\etocsettagdepth{mtappendix}{none}
\usepackage[breakable]{tcolorbox}
\usepackage{hyperref}

\newtcolorbox{boxL}{
    fontupper = \color{black},
    rounded corners,
    arc = 6pt,
    colframe = black!50, 
    boxrule = 0pt, 
    bottomrule = 4.5pt ,
    breakable,
}

\title{MARS: Multi-resolution Adaptive Routing \\for Sequential Recommendation}

\author{%
  Ming Yin\textsuperscript{1,2} \enspace
  Sixun Dong\textsuperscript{3} \enspace
  Yudong Liu\textsuperscript{1} \enspace
  Wen-Yun Yang\textsuperscript{2} \enspace
  Yunjiang Jiang\textsuperscript{2} \enspace
  Yiran Chen\textsuperscript{1} \\[0.7ex]
  {\normalfont\textsuperscript{1}Duke University \quad \textsuperscript{2}Meta \quad \textsuperscript{3} Independent}
}

\begin{document}

\maketitle

\begin{abstract}
Long-history recommenders often compress each user's history into a compact, candidate-independent memory that is cached and reused to score large candidate pools. We show that real user histories exhibit multi-scale semantic structure, with short-lived intent, medium-term interests, and long-term preferences coexisting in one sequence, and that monolithic cached memories preserve these scales unevenly: linear probes recover recent and mid-range content far worse than long-range content. We call this failure mode \textit{temporal aliasing}.
We propose \textbf{MARS}, a multi-resolution user memory that writes the full history into recurrent state tracks anchored to different half-lives, and a sparse routing reader that materializes compact seed memories by selecting the relevant temporal resolutions for each seed, preserving fixed-size candidate scoring.
MARS outperforms strong baselines on three public datasets, with gains that grow with history length.
Component-matched ablations with paired tests show that temporal diversity and selective routing each contribute beyond what hard-window memories or added capacity provide. The advantage of MARS over its interface-matched baseline also widens after within-user behavioral shifts, at about $1.02\times$ that baseline's warm-cache serving latency for $1{,}000$ candidates per user.
\end{abstract}

\section{Introduction}
\label{sec:intro}
Modern digital platforms rely on recommender systems~\citep{raza2025comprehensivereviewrecommendersystems} to optimize engagement metrics such as user retention, session duration, and click-through rate (CTR). Beyond collaborative filtering~\citep{item_cf} and matrix factorization~\citep{matrix_factorization}, transformer-based sequential models~\citep{liu2024mamba4recefficientsequentialrecommendation, sun2019bert4recsequentialrecommendationbidirectional} capture complex and long-range dependencies in user interaction histories.

Existing long-history architectures follow three paradigms. Full sequence modeling (e.g., HSTU~\citep{zhai2024actionsspeaklouderwords}) preserves complete histories, but its cost grows rapidly with history length, making it difficult to deploy under strict latency constraints~\citep{chai2025longerscalinglongsequence}. Target-specific sampling~\citep{qi2020searchbasedusermodelinglifelong, twin, twin_v2} retrieves candidate-aware subsequences, but truncates the history and requires candidate-dependent retrieval, so inference cost grows with the candidate pool. Summarization-based modeling, with VISTA~\citep{chen2026massivememorizationhundredstrillions} as a recent representative, compresses the full sequence into a fixed set of seed embeddings that are cached, refreshed incrementally or asynchronously, and reused across large candidate pools, bounding scoring cost independently of the raw history length.

Despite this efficient interface, existing summarization-based methods focus on compact compression and leave the temporal organization of reusable user memory largely unexplored. User behavior evolves at multiple temporal scales: short-lived browsing intent within a session, medium-term topical interests over days or weeks, and stable preferences that persist much longer. These signals coexist in one history while carrying different semantic content, yet without temporal-scale organization they are compressed into the same cached seed memory before the candidate scorer can access them. We refer to this as \textit{temporal aliasing}: temporally distinct behavioral evidence becomes entangled during memory construction, making it harder for downstream attention to recover the appropriate temporal signal. Band-wise probes make the effect measurable (Section~\ref{sec:temporal_probe}): from VISTA's cached memory, the content of the last $30$ interactions is recovered with $R^2=0.04$, versus $0.31$ for interactions at least $200$ steps old.

\begin{figure}[t]
    \centering
    \includegraphics[width=0.98\linewidth]{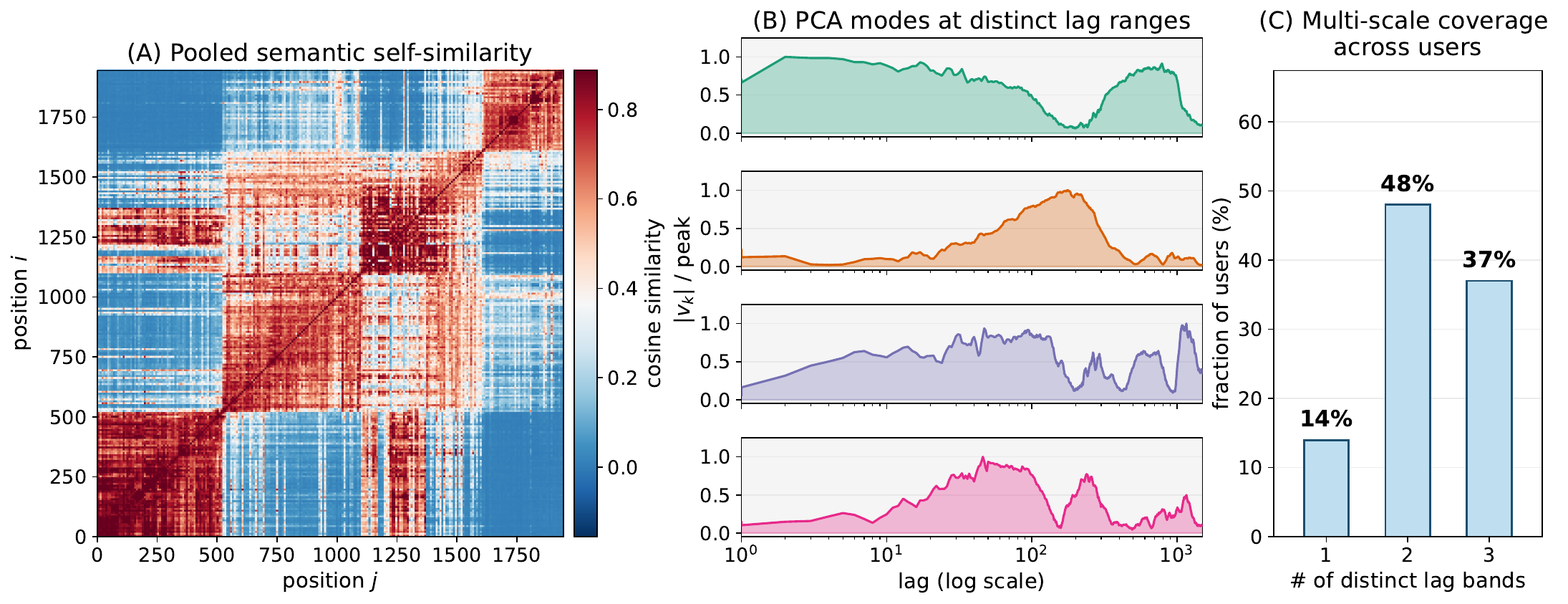}
    \caption{
\textbf{Multi-scale structure of user histories (KuaiRec).}
(A) Pooled semantic self-similarity for a long-history user shows coherent blocks at multiple temporal extents.
(B) PCA over lag-aligned similarity curves reveals dominant temporal loadings at distinct lag ranges. (C) For $85\%$ of long-history users, the top PCA modes span at least two coarse lag bands.
}
    \label{fig:motivation}
\end{figure}

To verify that multi-scale temporal structure is present in real behavior, we analyze KuaiRec~\citep{gao2022kuairec} and KuaiRand-1K~\citep{gao2022kuairand} using PPMI-SVD item representations~\citep{NIPS2014_b7866697,deerwester1990indexing}.
Figure~\ref{fig:motivation} shows coherent self-similarity blocks at multiple temporal extents, and the leading lag-aligned temporal modes span at least two coarse lag ranges for $85\%$ of long-history users; KuaiRand-1K shows consistent patterns (Appendix~\ref{app:kuairand_analysis}).

We propose \textbf{MARS} (\textbf{M}ulti-resolution \textbf{A}daptive \textbf{R}outing for \textbf{S}equential recommendation). MARS writes the full history into $M$ recurrent temporal tracks initialized with different half-life priors, forming a learnable temporal basis that ranges from rapid forgetting to long-term memory without hard window boundaries. A sparse seed-to-track router then materializes compact user memory: each seed reads a Top-$k$ subset of tracks, so the track bank provides temporal capacity while routing provides selectivity. MARS keeps the candidate-independent serving interface: the scorer attends only to the fixed-size seed memory and is left unchanged. Our scope is therefore long-history CTR-style recommendation with a cached user memory reused across a large candidate pool. Our main contributions are:
\begin{itemize}
  \setlength{\itemsep}{0.2em}
  \setlength{\topsep}{0.1em}
  \setlength{\parsep}{0pt}
  \setlength{\partopsep}{0pt}
    \item \textbf{Diagnosis of temporal aliasing.}
    We identify temporal aliasing as a structural limitation of monolithic cached memory for long histories, and support it with model-free evidence of multi-scale structure and band-wise recoverability probes.
    \item \textbf{Multi-resolution adaptive user memory.}
    We propose MARS, which organizes reusable user memory as half-life-anchored recurrent tracks read through sparse, history-conditioned routing, with the same scorer-facing cache and scoring cost as VISTA.
    \item \textbf{Controlled empirical validation.}
    MARS improves over strong baselines on three public datasets, with gains that grow with history length. Component-matched ablations with paired tests, window-based and capacity-scaled controls, a behavioral-shift evaluation, and measured A100 latency support the design choices and their serving practicality.
\end{itemize}

\section{Related Work}
\label{sec:related_work}

\paragraph{Sequential Recommendation.}
Sequential recommendation models~\citep{liu2024mamba4recefficientsequentialrecommendation,hidasi2016sessionbasedrecommendationsrecurrentneural} have progressed from self-attentive encoders such as SASRec~\citep{kang2018selfattentivesequentialrecommendation} and BERT4Rec~\citep{sun2019bert4recsequentialrecommendationbidirectional} to long-history systems under strict latency constraints.
Target-aware models such as DIN~\citep{zhou2018deepnetworkclickthroughrate,zhou2018deepevolutionnetworkclickthrough}, SIM~\citep{qi2020searchbasedusermodelinglifelong}, and TWIN~\citep{twin,twin_v2} retrieve or attend to candidate-specific histories, and MIRRN~\citep{xu2025multigranularityretrievalrefinementnetwork} refines retrieved evidence at multiple granularities; their target-conditioned history processing limits reuse across large candidate pools.
Full-sequence models~\citep{zhai2024actionsspeaklouderwords} preserve richer history at higher cost, whereas summarization-based methods such as VISTA~\citep{chen2026massivememorizationhundredstrillions} compress histories into reusable seed memories.
Multi-interest models~\citep{li2019mind,cen2020comirec} organize multiple user vectors by semantic interest, and user-memory networks~\citep{chen2018sequentialmemory,pi2019mimn} maintain explicit memory slots with read/write operations; MARS instead follows the candidate-independent cached-memory interface and organizes reusable memory by temporal retention scale.

\paragraph{Efficient Sequence Modeling.}
Linear attention~\citep{katharopoulos2020transformersrnnsfastautoregressive} and gated variants~\citep{yang2024gatedlinearattentiontransformers,sun2023retentivenetworksuccessortransformer} reduce quadratic cost via recurrent state summaries, and state space models~\citep{gu2022efficientlymodelinglongsequences} such as Mamba~\citep{gu2024mambalineartimesequencemodeling,dao2024transformersssmsgeneralizedmodels} and Hydra~\citep{hwang2024hydrabidirectionalstatespace} add input-dependent dynamics. Multi-timescale recurrent models parameterize hidden components with different temporal ranges~\citep{koutnik2014clockworkrnn,tallec2018recurrentneuralnetworkswarp}, and multi-scale SSMs~\citep{karadag2025msmamba} use multiple sampling rates, but these target sequence backbones rather than cached user memory.
MARS instead organizes gated rank-one recurrent states into half-life-anchored tracks and reads them into compact seed memories through sparse routing, making multi-timescale recurrence compatible with cached user memory (see also Appendix~\ref{app:related_work}).

\begin{figure}[t]
    \centering
    \includegraphics[width=\linewidth]{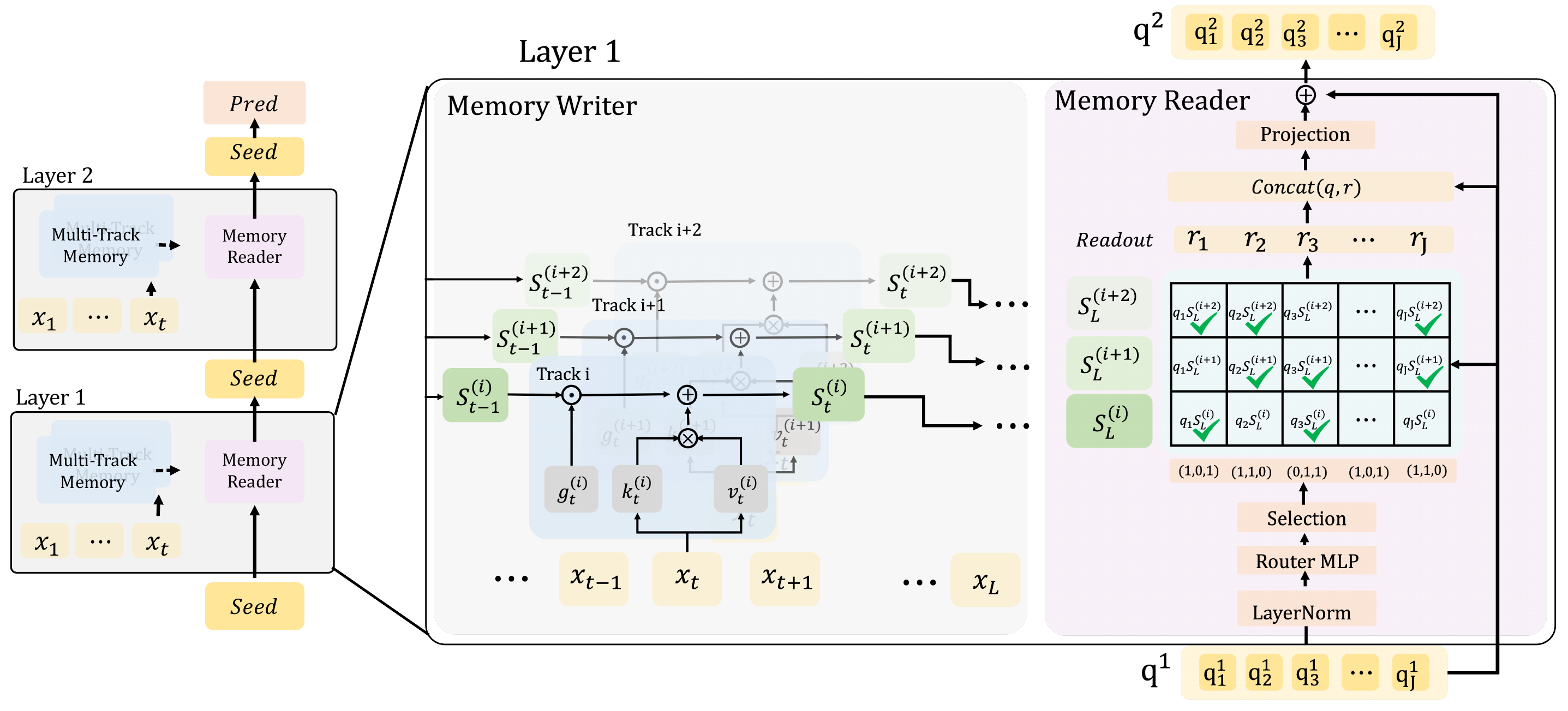}
   \caption{\textbf{MARS architecture.}
Each writer--reader layer writes the same history $\mathbf{x}_{1:L}$ into half-life-anchored recurrent tracks
$\mathbf{S}^{(i,\ell)}_L\in\mathbb{R}^{D\times D}$ with layer-specific parameters, and the reader routes each seed to a Top-$k$ subset of tracks (routing mask: rows are tracks, columns are seeds) for sparse readout and residual update.
Updated seeds feed the next layer, so deeper routers are history-conditioned.
The final seed memory $\mathbf{H}\in\mathbb{R}^{J\times D}$ is cached for target-conditioned prediction.}
    \label{fig:mars}
\end{figure}
\section{Method}
\label{sec:method}

We introduce MARS, a multi-resolution user memory for long-history recommendation with cached user representations.
Let $\mathbf{x}_{1:L}$ denote a user's interaction-history embeddings, and let $v$ be a candidate item with embedding $\mathbf{e}_v$.
MARS compresses $\mathbf{x}_{1:L}$ into compact seed memory while preserving behavior at different temporal scales. Formally, MARS decomposes prediction into
\begin{equation}
    \mathcal{S} = \operatorname{Write}(\mathbf{x}_{1:L}),
    \qquad
    \mathbf{H} = \operatorname{Read}(\mathcal{S}),
    \qquad
    \hat{y}_{v} = \operatorname{Score}(\mathbf{H},\mathbf{e}_v).
    \label{eq:mars_interface}
\end{equation}
MARS stacks $N_{\mathrm{mem}}$ writer--reader layers.
Each layer has its own writer parameters and processes the same history embeddings $\mathbf{x}_{1:L}$ into a layer-specific half-life-anchored multi-track state bank.
The layers are connected through seed representations updated by the reader, and the final-layer seeds form $\mathbf{H}\in\mathbb{R}^{J\times D}$.
Thus, long-history memory construction is separated from target-conditioned prediction: $\mathbf{H}$ is built once for the user and reused across candidate items.

\subsection{Memory Write: State Tracks with Half-Life Anchors}
\label{subsec:evolution}

We describe one memory layer and omit the layer index $\ell$.
In the full model, each memory layer maintains $M$ recurrent state tracks, where each track is anchored to a different retention scale.
The writer processes the full history $\mathbf{x}_{1:L}$ with every track, so tracks differ by their default memory span rather than by hard partitions of the sequence.

\paragraph{Recurrent state update.}
At step $t$, each track receives its own key $\mathbf{k}_t^{(i)}$ and value $\mathbf{v}_t^{(i)}$ by partitioning
$\mathbf{x}_t\mathbf{W}_{kv}$, where $\mathbf{W}_{kv}\in\mathbb{R}^{D\times 2MD}$ is a shared key-value projection.
The forget gate is dimension-wise and track-specific.
We compute all track gates jointly with a shared gate network
$\mathbf{F}_g:\mathbb{R}^{D}\rightarrow\mathbb{R}^{MD}$:
\begin{equation}
    [\mathbf{g}_t^{(1)},\ldots,\mathbf{g}_t^{(M)}]
    =
    \sigma\!\left(
        \mathbf{F}_g(\mathbf{x}_t)
        +
        [\mathbf{b}_g^{(1)},\ldots,\mathbf{b}_g^{(M)}]
    \right),
    \qquad
    \mathbf{g}_t^{(i)}\in\mathbb{R}^{D}.
    \label{eq:gate}
\end{equation}
Here $\mathbf{b}_g^{(i)}$ sets the base retention scale of track $i$, while the input-dependent offset from $\mathbf{F}_g(\mathbf{x}_t)$ allows the realized gate to adapt to the current interaction.

The state matrix $\mathbf{S}_t^{(i)}\in\mathbb{R}^{D\times D}$ then evolves by a gated rank-one recurrence:
\begin{equation}
  \mathbf{S}_t^{(i)}
=
(\mathbf{g}_t^{(i)} \mathbf{1}^{\top})
\odot
\mathbf{S}_{t-1}^{(i)}
+
\mathbf{k}_t^{(i)} (\mathbf{v}_t^{(i)})^{\top},
    \qquad
    \mathbf{S}_0^{(i)}=\mathbf{0}.
    \label{eq:state_update}
\end{equation}
This update gives each track a soft receptive field over the history and supports streaming inference through one rank-one update per track.

\paragraph{Half-life anchors.}
The base memory span of track $i$ is specified by a target half-life $H_i$, defined as the number of steps after which a past contribution decays to one half under the base gate.
For a constant base gate $\bar{g}_i$, the half-life condition
$\bar{g}_i^{H_i}=1/2$ gives $\bar{g}_i=2^{-1/H_i}$.
We therefore set the base gate bias as
\begin{equation}
    \mathbf{b}_g^{(i)}
    =
    \operatorname{logit}(2^{-1/H_i})\mathbf{1}_D.
    \label{eq:halflife_bias}
\end{equation}
where $\operatorname{logit}(p) \!=\! \log(p/(1-p))$ is the inverse of $\sigma$. The half-lives $\{H_i\}_{i=1}^{M}$ are logarithmically spaced between $H_{\min}$ and $H_{\max}$, forming a multi-resolution basis from fast to slow memory.
We keep $\mathbf{b}_g^{(i)}$ fixed during training to preserve a stable prior ordering of base retention scales.
The half-life anchor specifies only the default retention scale. Realized gates adapt through the input-dependent offset in \Eqref{eq:gate}, but only locally: on held-out KuaiRec users, the p05--p95 realized half-life of every track stays within $8.1\%$ of its anchor, and the fast-to-slow ordering is never violated (Table~\ref{tab:temporal_org}(a) and Appendix~\ref{app:realized_halflife}). After processing the history, the final states $\{\mathbf{S}_{L}^{(i)}\}_{i=1}^{M}$ form the track bank passed to the read module.

\subsection{Memory Read: Sparse Seed-to-Track Routing}
\label{subsec:routing}

The write module produces a bank of final state tracks
$\{\mathbf{S}_{L}^{(i,\ell)}\}_{i=1}^{M}$ at each memory layer $\ell$.
The read module converts this multi-resolution state bank into a fixed-size compact memory
$\mathbf{H}\in\mathbb{R}^{J\times D}$.
To do so, MARS maintains $J$ virtual seeds. The initial seed representations $\{\mathbf{q}_j^{(1)}\}_{j=1}^{J}\subset\mathbb{R}^{1\times D}$ are learnable parameters. At layer $\ell$, the seed $\mathbf{q}_j^{(\ell)}\in\mathbb{R}^{1\times D}$ acts as a memory query that selects a subset of tracks, reads from the selected states, and updates its representation.
\paragraph{Routed readout and residual update.}
The routing weights $\pi_{j,i}^{(\ell)}$, defined below, determine which tracks seed $j$ reads from at layer $\ell$.
Given the layer-$\ell$ track bank, seed $j$ performs a routed readout and residual update:
\begin{equation}
    \mathbf{r}_{j}^{(\ell)}
    =
    \sum_{i=1}^{M}
    \pi_{j,i}^{(\ell)}
    \left(
        \mathbf{q}_{j}^{(\ell)}
        \mathbf{S}_{L}^{(i,\ell)}
    \right),
    \qquad
    \mathbf{q}_{j}^{(\ell+1)}
    =
    \mathbf{q}_{j}^{(\ell)}
    +
    \operatorname{Out}^{(\ell)}\!\left(
        [\mathbf{q}_{j}^{(\ell)};\,\mathbf{r}_{j}^{(\ell)}]
    \right).
    \label{eq:seed_update}
\end{equation}
Here $[\cdot;\cdot]$ denotes row-wise concatenation.
The output block
$\operatorname{Out}^{(\ell)}:\mathbb{R}^{1\times 2D}\rightarrow\mathbb{R}^{1\times D}$
is layer-specific, applied independently to each seed, and shared across seeds within the layer.
In our implementation, it is a lightweight gated feedforward block.

\paragraph{Sparse track selection.}
The router maps each seed representation to $M$ track scores:
\begin{equation}
    \mathbf{z}_{j}^{(\ell)}
    =
    \operatorname{Router}^{(\ell)}(\mathbf{q}_{j}^{(\ell)})
    \in \mathbb{R}^{M},
\end{equation}
where $z_{j,i}^{(\ell)}$ is a seed-specific preference score over temporal tracks. During training, we add Gumbel noise $\boldsymbol{\epsilon}\sim\operatorname{Gumbel}(0,1)$ to encourage exploration over track assignments.
The forward routing weights are
\begin{equation}
    \pi_{j,i}^{(\ell)}
    =
    \frac{1}{k}
    \mathbf{1}\!\left[
        i\in\operatorname{TopK}
        \left(\mathbf{z}_{j}^{(\ell)}+\boldsymbol{\epsilon}\right)
    \right].
    \label{eq:topk_router}
\end{equation}
Thus, selected tracks receive equal weight $1/k$, and unselected tracks receive zero, preserving scale separation in the forward representation.
For backpropagation, we use a straight-through estimator: gradients flow through the dense softmax $\operatorname{Softmax}((\mathbf{z}_{j}^{(\ell)}+\boldsymbol{\epsilon})/\tau)$ over all $M$ tracks, while the forward routing weights remain those of \Eqref{eq:topk_router}.
The temperature $\tau$ controls the smoothness of this gradient path.
At inference, Gumbel noise is removed and deterministic Top-$k$ routing is used.

\paragraph{Hierarchical routing across layers.}
First-layer seeds do not depend on the history, so first-layer routing is seed-specific but history-independent.
Each routed readout injects history-specific information into the seeds through the residual update in \Eqref{eq:seed_update}, so deeper routing becomes history-conditioned, moving from static seed specialization to adaptive track selection.
The final-layer seeds form the compact memory $\mathbf{H}\in\mathbb{R}^{J\times D}$.

\subsection{Target-Conditioned Scoring}
\label{subsec:scoring}

Given the compact memory $\mathbf{H}$ and a candidate item embedding $\mathbf{e}_v$, the scorer attends from the candidate to the cached seed memory and predicts the response $\hat{y}_v = f_{\mathrm{score}}(\mathbf{e}_v, \mathbf{H})$, where $f_{\mathrm{score}}$ is instantiated as standard target-aware attention followed by a prediction head. Since the scorer only reads $\mathbf{H}\in\mathbb{R}^{J\times D}$, its cost is independent of the history length $L$, and the same $\mathbf{H}$ is reused to score all candidates of the user.
Across summarization-based variants, we keep $f_{\mathrm{score}}$ fixed so that empirical differences isolate the effect of memory construction.

\subsection{Training and Inference}
\label{subsec:training_inference}

Following standard practice, MARS is trained for CTR prediction with binary cross-entropy.

\paragraph{Parallel training.}
Although \Eqref{eq:state_update} is recurrent, it can be parallelized using the gated linear attention scan~\citep{yang2024gatedlinearattentiontransformers}.
The key observation is that cumulative gate products can be rewritten as differences of prefix sums in log-space:
\begin{equation}
    \log\prod_{n=m+1}^{t} \mathbf{g}_n^{(i)}
    =
    \operatorname{CumSum}(\log\mathbf{g}^{(i)})_t
    -
    \operatorname{CumSum}(\log\mathbf{g}^{(i)})_m.
\end{equation}
Details of the parallel scan formulation are provided in Appendix~\ref{app:parallel_scan}.
This enables fused chunk-wise kernels with linear work in the sequence length $L$.

\paragraph{Serving.}
MARS caches the compact memory $\mathbf{H}$, which is all the candidate scorer reads; under periodic or asynchronous refresh, nothing else is stored.
When event-level freshness is required, MARS can additionally cache the state bank
$\mathcal{S}=\{\mathbf{S}_{L}^{(i,\ell)}\}_{i,\ell}$: a new interaction $\mathbf{x}_{L+1}$ then updates each track in each memory layer by one rank-one step, and $\mathbf{H}$ is re-materialized once by sparse readout and reused for all candidate items until the next update.

\section{Experiments}
\label{sec:exp}

\subsection{Experimental Setup}
\label{subsec:setup}
\textbf{Datasets.} We evaluate MARS on three public datasets: Amazon-Electronics\footnote{\url{https://github.com/reczoo/Datasets/tree/main/Amazon/AmazonElectronics_x1}}, KuaiRand-1K~\citep{gao2022kuairand}, and KuaiRec~\citep{gao2022kuairec}. All datasets are formulated as CTR prediction tasks.

\begin{wraptable}{r}{0.5\linewidth}
\vspace{-1.0em}
\centering
\caption{Dataset statistics after pre-processing.}
\label{tab:datasets}
\vspace{-0.2em}
\fontsize{8.8pt}{10pt}\selectfont
\setlength{\tabcolsep}{4pt}
\renewcommand{\arraystretch}{1.0}
\begin{tabular}{lrrr}
\toprule
\textbf{Dataset} & \textbf{Samples} & \textbf{Avg. Len} & \textbf{Max Len} \\
\midrule
Amazon-Elec.  & 1.69M  & 8.9   & 429 \\
KuaiRand-1K   & 4.02M  & 957   & 1,000 \\
KuaiRec     & 12.53M & 1,721 & 3,000 \\
\bottomrule
\end{tabular}
\vspace{-0.5em}
\end{wraptable}
For controlled comparisons, we use the same maximum history length for all methods and retain the most recent interactions whenever a user history exceeds this budget. This protocol keeps quadratic full-attention baselines computationally feasible while ensuring that every method receives the same input context. For KuaiRand-1K and KuaiRec, the history budgets are set to $1$k and $3$k, which remain sufficiently long to evaluate the long-history regime. Dataset statistics are summarized in Table~\ref{tab:datasets}, and additional details are provided in Appendix~\ref{app:public_protocol}.

\textbf{Baselines and metrics.}
We compare with target-aware CTR models (DIN~\citep{zhou2018deepnetworkclickthroughrate}, DIEN~\citep{zhou2018deepevolutionnetworkclickthrough}), a two-tower model (TTSN~\citep{covington2016ttsn}), attention-based sequence models (standard Multi-Head Attention (MHA)~\citep{vaswani2017attention}, SASRec~\citep{kang2018selfattentivesequentialrecommendation}), efficient long-sequence backbones (RetNet~\citep{sun2023retentivenetworksuccessortransformer}, Mamba4Rec~\citep{liu2024mamba4recefficientsequentialrecommendation}), HSTU~\citep{zhai2024actionsspeaklouderwords}, and the summarization-based VISTA~\citep{chen2026massivememorizationhundredstrillions}, which shares the cached-memory interface of MARS and is our primary interface-matched baseline (Appendix~\ref{app:serving_interfaces}). All methods share the same splits and CTR pipeline~\citep{chen2026massivememorizationhundredstrillions}; we report AUC and normalized entropy (NE)~\citep{he2014practical} as mean $\pm$ std over three runs.

\subsection{Overall Performance}
\label{subsec:overall}

\newcommand{\mstd}[2]{\ensuremath{#1{\scriptstyle\,\pm\,#2}}}
\newcommand{\bmstd}[2]{\ensuremath{\mathbf{#1}{\scriptstyle\,\pm\,#2}}}

\begin{table}[t]
\centering
\caption{Overall performance with mean $\pm$ std over 3 runs. Higher AUC and lower NE are better. Comparisons between MARS and the interface-matched VISTA baseline are significant under paired user-level bootstrap tests, with $p<0.05$ on Amazon and $p<0.001$ on KuaiRand and KuaiRec.
}
\label{tab:overall}
\vspace{0.3em}
\fontsize{9.5pt}{11pt}\selectfont
\setlength{\tabcolsep}{3.5pt}
\renewcommand{\arraystretch}{1.0}
\hspace*{-0.15in}
\begin{tabular}{l cc cc cc}
\toprule
& \multicolumn{2}{c}{\textbf{Amazon}} 
& \multicolumn{2}{c}{\textbf{KuaiRand-1K}} 
& \multicolumn{2}{c}{\textbf{KuaiRec}} \\
\cmidrule(lr){2-3} \cmidrule(lr){4-5} \cmidrule(lr){6-7}
\textbf{Method} 
& AUC $\uparrow$ & NE $\downarrow$ 
& AUC $\uparrow$ & NE $\downarrow$ 
& AUC $\uparrow$ & NE $\downarrow$ \\
\midrule
DIN        & \mstd{0.873}{8e{-}4} & \mstd{0.656}{1e{-}4} & \mstd{0.744}{0.003} & \mstd{0.864}{0.005} & \mstd{0.779}{0.002} & \mstd{0.852}{0.002} \\
DIEN       & \mstd{0.874}{0.001}  & \mstd{0.651}{0.005}  & \mstd{0.742}{0.001} & \mstd{0.867}{0.002} & \mstd{0.781}{0.001} & \mstd{0.849}{0.004} \\
TTSN       & \mstd{0.877}{0.005}  & \mstd{0.644}{0.010}  & \mstd{0.740}{0.003} & \mstd{0.869}{0.004} & \mstd{0.784}{0.001} & \mstd{0.856}{0.003} \\
MHA        & \mstd{0.881}{1e{-}4} & \mstd{0.634}{0.002}  & \mstd{0.743}{0.001} & \mstd{0.863}{0.005} & \mstd{0.776}{0.002} & \mstd{0.853}{5e{-}4} \\
SASRec     & \mstd{0.884}{4e{-}4} & \mstd{0.627}{0.001}  & \mstd{0.742}{0.003} & \mstd{0.868}{0.007} & 
\mstd{0.764}{2e{-}4} & \mstd{0.857}{0.001} \\
RetNet     & \mstd{0.883}{0.001} & \mstd{0.630}{0.002}  & \mstd{0.740}{5e{-}4} & \mstd{0.869}{0.002} & 
\mstd{0.781}{0.001} & \mstd{0.849}{0.002} \\
Mamba4Rec  & \mstd{0.883}{0.003}  & \mstd{0.632}{0.004}  & \mstd{0.739}{0.002} & \mstd{0.871}{0.003} & \mstd{0.785}{0.002} & \mstd{0.840}{0.004} \\
HSTU       & \mstd{0.884}{0.001}  & \mstd{0.628}{0.001}  & \mstd{0.743}{0.001} & \mstd{0.863}{1e{-}5} & \mstd{0.801}{0.001} & \mstd{0.795}{0.002} \\
VISTA      & \mstd{0.886}{0.002}  & \mstd{0.621}{0.003}  & \mstd{0.744}{0.001} & \mstd{0.863}{0.003} & \mstd{0.793}{0.002} & \mstd{0.819}{0.003} \\
\midrule
\textbf{MARS} 
           & \bmstd{0.887}{0.001} & \bmstd{0.618}{0.002} & \bmstd{0.748}{0.001} & \bmstd{0.857}{0.002} & \bmstd{0.804}{0.002} & \bmstd{0.791}{0.003} \\
\bottomrule
\end{tabular}
\end{table}

MARS improves over all baselines on the three datasets (Table~\ref{tab:overall}), and prior works~\citep{zhou2018deepnetworkclickthroughrate, zhou2018deepevolutionnetworkclickthrough} suggest that such CTR improvements are practically meaningful. On Amazon-Electronics, where histories are short, MARS is comparable to VISTA with slightly better mean AUC and NE; on KuaiRand-1K it obtains the best AUC and NE; and the largest gain appears on KuaiRec, which has the longest histories, where AUC improves from $0.793$ to $0.804$. We additionally evaluate MARS on an industrial recommendation dataset, where it consistently reduces evaluation NE compared with VISTA across all four tasks and two history-processing settings (single run per setting; Appendix~\ref{app:industrial_validation}).

\subsection{Analysis}

\subsubsection{Temporal-Scale Recoverability from Compressed Memory}
\label{sec:temporal_probe}
\begin{figure}[t]
    \centering
    \includegraphics[width=\textwidth]{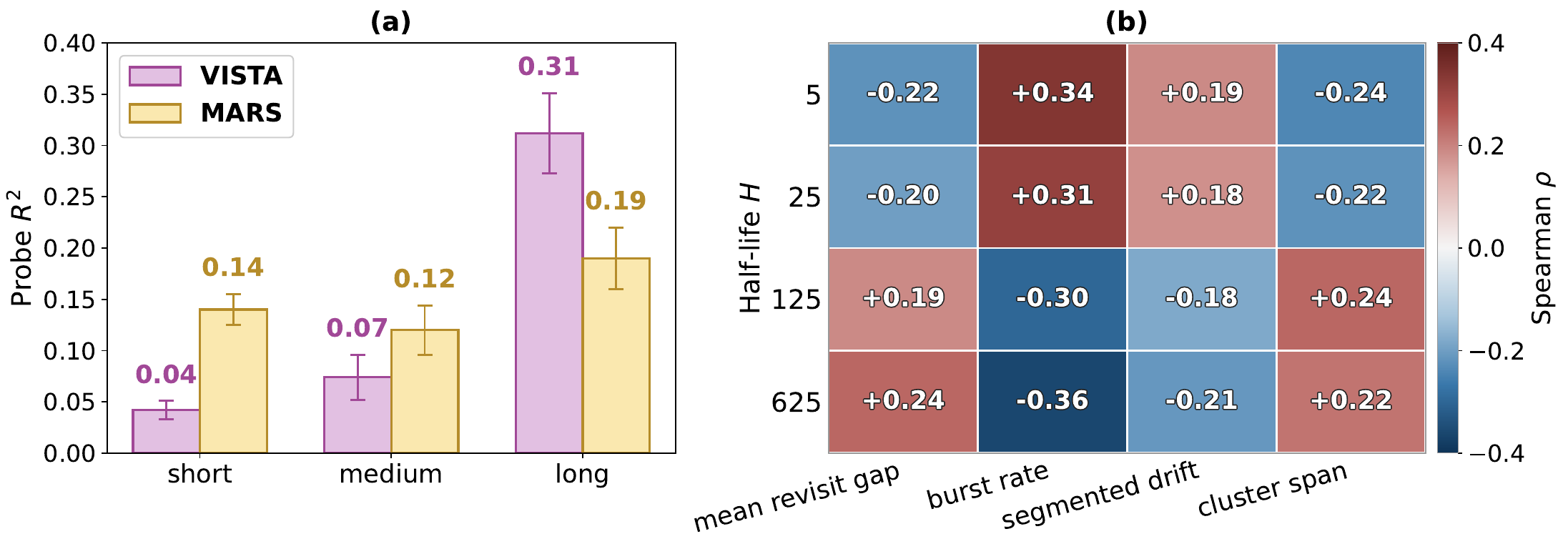}
    \caption{
    \textbf{Representation and routing evidence for multi-resolution memory.}
    (a) Linear probes: MARS reduces VISTA's long-band bias (error bars: std over 5 folds).
    (b) Spearman correlation between track usage and behavioral signals: short-half-life tracks align with bursty or drifting behavior and long-half-life tracks with persistent interests (all $q<0.01$, Benjamini--Hochberg).
    }
    \label{fig:temporal_routing_analysis}
\end{figure}

For $512$ long-history KuaiRec users, we build model-free item embeddings with PPMI-SVD and split each history into three recency bands: \textit{short} (last $30$ items), \textit{medium} (lag $30$--$200$), and \textit{long} (lag $\geq 200$), with each band's mean item embedding as the target.
Given each model's compressed memory $\mathbf{H}_u$ before candidate scoring, we apply PCA within each training fold and fit a Ridge probe per band, reporting 5-fold held-out $R^2$; a shuffled-memory baseline gives $R^2 \approx -0.03$.
Figure~\ref{fig:temporal_routing_analysis}(a) shows that VISTA is biased toward long-range content, with short/medium/long $R^2$ of $0.04/0.07/0.31$, whereas MARS yields a more balanced $0.14/0.12/0.19$ profile, reducing the cross-band range from $0.27$ to $0.07$ at a similar mean $R^2$.
A hard fixed-window memory (Section~\ref{subsec:ablation}) only partially reduces this bias (Table~\ref{tab:temporal_org}(b)); under a paired user-level bootstrap, MARS has a higher mean $R^2$ ($+0.022$, 95\% CI $[0.010, 0.034]$) and a smaller cross-band range ($0.07$ vs.\ $0.16$; CI of the difference $[0.05, 0.13]$) than this memory (Appendix~\ref{app:component_details}). The tracks themselves also keep their intended timescales: the realized half-lives stay in disjoint bands close to their anchors (Table~\ref{tab:temporal_org}(a)).
MARS thus restores information that monolithic or hard-window compression suppresses at shorter scales, consistent with reduced temporal aliasing.

\begin{table}[t]
\centering
\caption{\textbf{Temporal organization of the memory on KuaiRec.}
(a) Realized half-lives of the learned forget gates on $1{,}280$ held-out users; each range spans the minimum p05 to the maximum p95 across the three memory layers (Appendix~\ref{app:realized_halflife}).
(b) Held-out $R^2$ of linear probes predicting history-band centroids from the cached memory, rounded to two decimals; range is the maximum minus the minimum across bands (Appendix~\ref{app:component_details}).}
\label{tab:temporal_org}
\begin{minipage}[t]{0.40\textwidth}
\centering
\textbf{(a) Realized half-lives}\\[0.3em]
\small
\setlength{\tabcolsep}{2.5pt}
\label{tab:realized_halflife}
\begin{tabular}{ccc}
\toprule
Anchor $H_i$ & p05--p95 range & Max.\ rel.\ dev.\ \\
\midrule
$5$ & $5.0$--$5.0$ & $0.000$ \\
$25$ & $24.0$--$26.8$ & $0.072$ \\
$125$ & $120.6$--$124.4$ & $0.035$ \\
$625$ & $574.1$--$650.5$ & $0.081$ \\
\bottomrule
\end{tabular}
\end{minipage}
\hfill
\begin{minipage}[t]{0.57\textwidth}
\centering
\textbf{(b) Temporal recoverability}\\[0.3em]
\small
\setlength{\tabcolsep}{3.5pt}
\label{tab:fixed_window_probe}
\begin{tabular}{lccccc}
\toprule
Memory & Short & Medium & Long & Mean & Range $\downarrow$ \\
\midrule
VISTA & $0.04$ & $0.07$ & $\mathbf{0.31}$ & $0.14$ & $0.27$ \\
Fixed-window & $0.06$ & $0.10$ & $0.22$ & $0.13$ & $0.16$ \\
MARS & $\mathbf{0.14}$ & $\mathbf{0.12}$ & $0.19$ & $\mathbf{0.15}$ & $\mathbf{0.07}$ \\
\bottomrule
\end{tabular}
\end{minipage}
\end{table}

\subsubsection{User-Adaptive Routing Aligns with Semantic Signals}
\label{subsec:routing_semantics}
We define the usage of a temporal track for a user as the fraction of seeds whose Top-$k$ routing selects it, and correlate usage with user-level behavioral signals: mean revisit gap and cluster span capture longer-range persistence, while burst rate and segmented drift capture short-term concentration or local temporal instability.
Figure~\ref{fig:temporal_routing_analysis}(b) shows that short-half-life tracks ($H=5,25$) correlate positively with burst rate and segmented drift and negatively with mean revisit gap and cluster span, whereas long-half-life tracks ($H=125,625$) show the opposite trend. Because every seed selects $k$ tracks, per-user usages sum to $k$, so the short- and long-track rows mirror each other by construction; the informative part is which behavioral signals align with which retention scales.
Routing also depends partly on history length, but it does not collapse to a length-to-scale rule: short-half-life tracks remain active for many long-history users, and users with similar history lengths route differently (Appendix~\ref{app:routing_length}).
Nor does routing collapse: the scorer-facing layer uses $3.92$ of $4$ tracks effectively, and routing grows more user-specific with depth (Appendix~\ref{app:routing_diagnostics}).

\subsubsection{Routing and Robustness Under Behavioral Shift}
\label{subsec:shift}
We test whether history-conditioned routing helps when user behavior changes (Appendix~\ref{app:shift}).
\textit{Routing response.} For $4{,}096$ held-out long-history users, we compare final-layer routing masks computed with and without the most recent $30$ interactions. Masks are stable overall (Jaccard similarity $0.934$), but users in the highest drift tercile change their masks more than those in the lowest (Jaccard $0.931$ vs.\ $0.942$) and shift more capacity toward short-half-life tracks ($+0.055$ vs.\ $+0.047$ in short-track share; both differences $p<0.001$).
\textit{Robustness.} For $4{,}000$ users, we locate one shift point per user with a model-independent drift score and evaluate trained checkpoints, without retraining, on the $20$ interactions before and after it (Table~\ref{tab:shift}).
The AUC advantage of MARS over VISTA nearly triples after the shift. Against an independently trained user-independent Top-$k$ control with the same tracks, capacity, and routing budget, the advantage is inconclusive before the shift but significant after it, attributing part of the post-shift gain to history-conditioned routing. We scope this conclusion to within-user behavioral shifts rather than arbitrary distribution shifts.

\begin{table}[t]
\centering
\caption{\textbf{Paired advantage $A$ of MARS before and after within-user behavioral shifts}, and its change $G=A(\mathrm{post})-A(\mathrm{pre})$. Positive $\Delta$AUC and negative $\Delta$NE favor MARS. Brackets: 95\% CIs from $10{,}000$ paired user-level bootstrap resamples ($p<0.001$ for all gaps).}
\label{tab:shift}
\small
\setlength{\tabcolsep}{5pt}
\begin{tabular}{llccc}
\toprule
Baseline & Metric & Pre-shift $A$ & Post-shift $A$ & Gap $G$ [95\% CI] \\
\midrule
VISTA & $\Delta$AUC & $+0.006$ & $+0.017$ & $+0.011$ $[+0.006, +0.016]$ \\
 & $\Delta$NE & $-0.008$ & $-0.027$ & $-0.019$ $[-0.028, -0.010]$ \\
\midrule
User-independent Top-$k$ & $\Delta$AUC & $+0.0004$ & $+0.0036$ & $+0.0032$ $[+0.0020, +0.0044]$ \\
 & $\Delta$NE & $-0.0008$ & $-0.0056$ & $-0.0048$ $[-0.0065, -0.0031]$ \\
\bottomrule
\end{tabular}
\end{table}

\subsubsection{MARS Gains Grow with History Length}
\label{subsubsec:user_bin}
\begin{figure}[t]
    \centering
    \includegraphics[width=\textwidth]{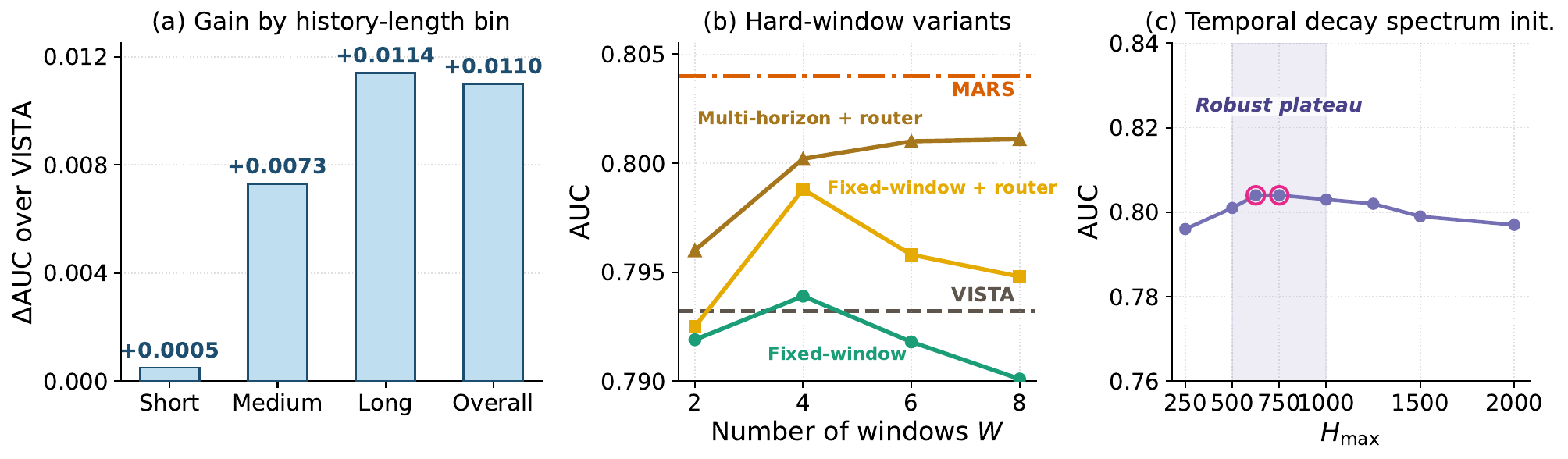}
    \caption{
    \textbf{Analysis and ablations on KuaiRec.}
    (a) MARS gains grow with history length.
    (b) Window-based variants improve when equipped with the MARS router, and multi-horizon windows beat disjoint fixed windows, but all remain below MARS.
    (c) A balanced half-life spectrum performs best, with the default $H_{\max}$ in a stable high-performing region.
    }
    \label{fig:analysis_all}
\end{figure}

Splitting the KuaiRec test set by effective history length $S$ into short ($S<500$), medium ($500 \leq S < 1500$), and long ($S\geq1500$) bins, the AUC gain of MARS over VISTA grows from $+0.0005$ to $+0.0073$ and $+0.0114$ (Figure~\ref{fig:analysis_all}(a)). Temporal-scale organization thus matters most when histories are long and contain heterogeneous behavioral evidence.

\subsection{Ablation Study}
\label{subsec:ablation}

All ablations use KuaiRec, whose long histories make it the most diagnostic setting for temporal resolution.

\begin{table}[t]
\centering
\caption{
\textbf{Ablations on KuaiRec.}
(a) Component decomposition (Top-$k$ readouts use $k=2$); a single-resolution bank has $M=4$ tracks sharing one retention scale. $^{\dagger}$: MARS is better on both AUC and NE, with 95\% paired user-level bootstrap CIs excluding zero (Appendix~\ref{app:component_details}).
(b) Sensitivity to the number of tracks $M$ and routing budget $k$ (AUC / NE). $^{*}$With $k=M$, every track is selected, and our implementation then weights the tracks by the router softmax, so these entries correspond to dense softmax readout rather than uniform averaging.
}
\label{tab:ablation_summary}
\begin{minipage}[t]{0.51\textwidth}
\centering
\textbf{(a) Component decomposition}\\[0.3em]
\footnotesize
\setlength{\tabcolsep}{3.5pt}
\renewcommand{\arraystretch}{1.05}
\begin{tabular}{lcc}
\toprule
Track bank, readout & AUC $\uparrow$ & NE $\downarrow$ \\
\midrule
Single track ($M{=}1$)$^{\dagger}$ & $0.794$ & $0.812$ \\
Single-res.\ bank, uniform & $0.795$ & $0.809$ \\
Single-res.\ bank, Top-$k$$^{\dagger}$ & $0.797$ & $0.804$ \\
Multi-res.\ bank, uniform$^{\dagger}$ & $0.797$ & $0.806$ \\
Multi-res.\ bank, dense softmax & $0.802$ & $0.795$ \\
Multi-res.\ bank, Top-$k$ + softmax & $0.802$ & $0.794$ \\
Multi-res.\ bank, user-indep.\ Top-$k$ & $0.799$ & $0.801$ \\
\midrule
\textbf{MARS} (multi-res.\ bank, Top-$k$) & $\mathbf{0.804}$ & $\mathbf{0.791}$ \\
\bottomrule
\end{tabular}
\end{minipage}
\hfill
\begin{minipage}[t]{0.47\textwidth}
\centering
\textbf{(b) Sensitivity to $M$ and $k$}\\[0.3em]
\footnotesize
\setlength{\tabcolsep}{1.8pt}
\renewcommand{\arraystretch}{1.05}
\begin{tabular}{c|ccc}
\toprule
$M \backslash k$ & $1$ & $2$ & $4$ \\
\midrule
$1$ & 0.794 / 0.812 & --- & --- \\
$2$ & 0.798 / 0.803 & 0.799 / 0.797$^{*}$ &  ---\\
$4$ & 0.801 / 0.795 & \textbf{0.804 / 0.791} & 0.802 / 0.795$^{*}$ \\
$8$ & 0.798 / 0.806 & 0.802 / 0.797 & 0.801 / 0.798 \\
\bottomrule
\end{tabular}
\end{minipage}
\end{table}

\paragraph{Component decomposition.}
Table~\ref{tab:ablation_summary}(a) separates track capacity, temporal diversity, and readout. A \textit{single-resolution bank}, whose $M=4$ tracks share one retention scale but keep independent projections, preserves capacity and nontrivial routing while removing temporal diversity: it barely improves over a single track, and routing over it adds only $+0.002$ AUC and $-0.005$ NE. Over the multi-resolution bank, the same routing adds $+0.007$ AUC and $-0.015$ NE relative to uniform readout, so temporal diversity and selective routing are complementary. MARS beats the capacity-matched single-resolution Top-$k$ control by $0.007$ AUC and $0.013$ NE, and its gaps to this control, to uniform readout, and to a single track are significant under paired bootstrap tests (Appendix~\ref{app:component_details}).
Removing history conditioning (routing that depends only on the seed index) or sparsity (dense softmax) also hurts, and softmax weighting within the selected set trails uniform mixing, likely by over-concentrating on the top-scored track.
Finally, a constant-anchor control without the input-dependent gate offset ($\mathbf{g}_t=\sigma(\mathbf{b}_g)$) lowers AUC by $0.0009$ (95\% CI $[0.0003, 0.0015]$) and raises NE by $0.0022$ ($[0.0010, 0.0034]$): adaptation beyond the fixed anchors is modest but significant, and the larger adaptive benefit comes from history-conditioned composition across tracks.

\paragraph{Sensitivity to $M$ and $k$.}
In Table~\ref{tab:ablation_summary}(b), increasing $M$ from $1$ to $4$ improves AUC from $0.794$ to $0.801$ under $k=1$, and $M=4$, $k=2$ performs best. At $M=4$, AUC varies by only $0.003$ across $k$: $k=1$ restricts each seed to one resolution, whereas $k=4$ selects every track and, in our implementation, reduces to dense softmax readout ($0.802/0.795$, in line with the dense-softmax variant in Table~\ref{tab:ablation_summary}(a)). $M=8$ brings no further gain. Sensitivity to $J$ and $N_{\mathrm{mem}}$ is in Appendix~\ref{app:hyper_sensitivity}.

\paragraph{Window-based variants.}
Under the same scorer, we test $W$ disjoint fixed windows with uniform aggregation, the same windows read by the MARS router, and $W$ overlapping suffix horizons read by the router (Figure~\ref{fig:analysis_all}(b)). Routing helps and multi-horizon windows beat disjoint ones, but all variants remain below MARS and saturate or degrade as $W$ grows. A simpler baseline that concatenates mean-pooled summaries over suffix windows of $\{30, 200, 1000, 2000\}$ interactions also improves over VISTA ($0.797\pm0.001$ vs.\ $0.793\pm0.002$ AUC, three seeds) but stays below MARS ($0.804\pm0.002$). Hard, index-based cutoffs thus recover part of the aliased signal, while continuous half-life retention with selective readout recovers more.

\paragraph{Half-life spectrum.}
Fixing $H_{\min}$ and varying $H_{\max}$ (Figure~\ref{fig:analysis_all}(c)), performance drops when $H_{\max}$ is too small, leaving insufficient long-range retention, and also degrades when it is very large, likely because overly persistent tracks mix in stale evidence. The default lies in a broad high-performing region: MARS benefits from a temporally diverse but bounded spectrum.

\subsection{Efficiency Analysis}
\label{sec:efficiency}

In warm-cache serving, the compact memory $\mathbf{H}_u\in\mathbb{R}^{J\times D}$ is materialized once per user refresh and reused for all candidates. VISTA and MARS expose the same $\mathbf{H}_u$ to the same scorer, so per-candidate scoring performs the same computation, independent of the history length $L$; they differ only on the update path, where one incremental MARS update costs $O(N_{\mathrm{mem}}MD^2)$ for the tracks and $O(N_{\mathrm{mem}}JkD^2)$ for sparse re-materialization.

\paragraph{Cache footprint.}
In FP32, $\mathbf{H}_u$ takes $16$ KiB per user for both methods, and MARS stores nothing else under periodic refresh. The recurrent state bank ($48$ KiB) is needed only for event-level incremental updates without history replay, and it compresses well: diagonal or projected rank-$4$ states bring the total cache to $17.5$ or $16.75$ KiB while keeping most of the gain (single-run AUC $0.8012$ and $0.8003$, vs.\ $0.8041$ with full states and $0.7932$ for VISTA). It can also be kept only for recently active users (Appendices~\ref{app:compressed_state_cache} and~\ref{app:cache_footprint}).

\paragraph{Measured latency.}
On one NVIDIA A100 (batches of $256$ users, medians over $40$ runs), an incremental update plus seed-memory materialization takes $0.421$--$0.425$ ms per batch as the history length grows from $500$ to $5{,}000$, so warm-cache serving does not replay the raw history. With $1{,}000$ candidates per user, a practical ranking-stage workload~\citep{li2024contextual,song2022rethinking}, end-to-end latency is $15.48$ ms for MARS versus $15.15$ ms for VISTA (about $1.02\times$ at both the median and p95), i.e., $16.5$M candidate evaluations per second; the scoring-only latency ratio is $1.009$--$1.014\times$ (Appendix~\ref{app:latency}). These measurements exclude one-time initialization from the raw history.

\begin{figure}[t]
\centering
\newlength{\panelheight}
\setlength{\panelheight}{1.45in}
\begin{minipage}[t]{0.46\textwidth}
\vspace{0pt}
\centering
\begin{minipage}[t][\panelheight][c]{\linewidth}
\centering
\small
\setlength{\tabcolsep}{6pt}
\renewcommand{\arraystretch}{1.15}
\resizebox{0.96\linewidth}{!}{
\begin{tabular}{lcc}
\toprule
Config & Update FLOPs / event & AUC $\uparrow$ \\
\midrule
VISTA-base        & $3.33$M                  & 0.7932 \\
VISTA-4heads      & $6.87$M ($2.1{\times}$)  & 0.7939 \\
VISTA-wider       & $12.98$M ($3.9{\times}$) & 0.7943 \\
VISTA-deeper      & $8.23$M ($2.5{\times}$)  & 0.7949 \\
\textbf{MARS}     & $3.45$M ($1.04{\times}$) & \textbf{0.8041} \\
\bottomrule
\end{tabular}
}
\end{minipage}

\vspace{0.35em}
\captionof{table}{
\textbf{Warm-cache memory-update cost} on KuaiRec.
FLOPs for updating the compact user memory after a new interaction, relative to VISTA-base.
}
\label{tab:warm_cache_streaming_cost}
\end{minipage}
\hfill
\begin{minipage}[t]{0.5\textwidth}
\vspace{0pt}
\centering
\begin{minipage}[t][\panelheight][c]{\linewidth}
\centering
\includegraphics[height=\panelheight,width=\linewidth,keepaspectratio]{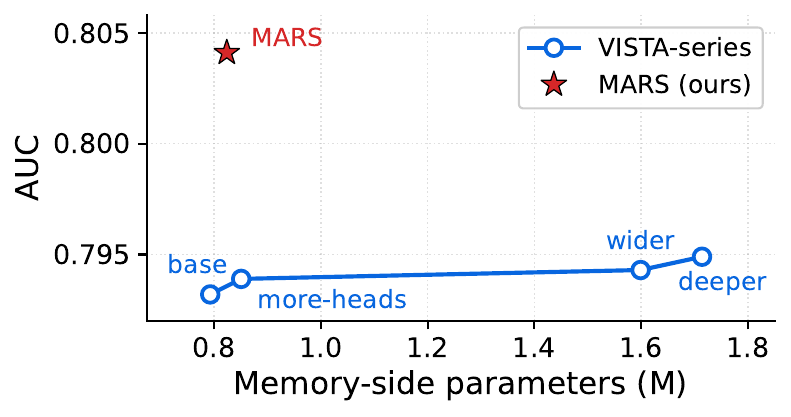}
\end{minipage}

\vspace{0.35em}
\captionof{figure}{
\textbf{Capacity-controlled comparison.}
MARS versus scaled VISTA variants; the x-axis reports memory-side parameters, excluding shared item embeddings.
}
\label{fig:vista_scaling_curve}
\end{minipage}
\end{figure}

\paragraph{Capacity-controlled comparison.}
MARS needs $3.45$M update FLOPs per event, $1.04\times$ VISTA-base (Table~\ref{tab:warm_cache_streaming_cost}). Since this update is materialized once per user refresh, the extra computation is not multiplied by the candidate-pool size. We also test whether the gain can be explained by giving VISTA more memory-side capacity: VISTA-4heads adds parallel attention heads, VISTA-wider doubles the hidden dimension $D$, and VISTA-deeper doubles the number of memory-construction layers $N_{\mathrm{mem}}$. These variants raise AUC only from $0.7932$ to at most $0.7949$, versus $0.8041$ for MARS, even at up to $3.9\times$ the update FLOPs (Table~\ref{tab:warm_cache_streaming_cost} and Figure~\ref{fig:vista_scaling_curve}). Capacity and compute alone therefore do not explain the gain; the difference reflects the structural change introduced in Section~\ref{sec:method}.

\section{Conclusion}
\label{sec:conclusion}
We presented MARS, a multi-resolution user memory for long-history recommendation with cached user representations, which writes full histories into half-life-anchored recurrent tracks and reads them through sparse, history-conditioned seed-to-track routing.
MARS reduces temporal aliasing, improves over strong baselines with gains that grow with history length, and widens its advantage after within-user behavioral shifts, at a warm-cache serving cost close to that of VISTA.
Our evidence is offline and scoped to CTR-style prediction with cached memory (limitations in Appendix~\ref{sec:limitations}).
These findings highlight temporal organization as a useful design principle for reusable user memory.

\begin{ack}
This work was supported by the National Science Foundation under Grant No.~2625580 and by the Army Research Office under Grant No.~W911NF-23-2-0224.
\end{ack}

\newpage
\bibliography{reference}
\bibliographystyle{plainnat}

\appendix
\clearpage
\onecolumn
\appendix

\section{Additional Multi-scale Structure Analysis}
\label{app:kuairand_analysis}

This appendix provides details for the co-occurrence analysis used in
Figure~\ref{fig:motivation}. The goal of this analysis is to test whether real
user histories contain semantic structure at multiple temporal scales, independent
of any learned recommendation model.

\paragraph{Model-free item representations.}
For each dataset, we construct item embeddings from user histories using
PPMI co-occurrence followed by truncated SVD. Specifically, we build an item
co-occurrence matrix from training histories using a context window of 100
interactions, convert it to a positive pointwise mutual information (PPMI)
matrix, and apply truncated SVD to obtain 64-dimensional item vectors. We
$\ell_2$-normalize the resulting item embeddings before computing similarities.
Because these embeddings are derived only from item co-occurrence statistics,
they provide a model-free semantic space for analyzing historical behavior.

\paragraph{Lag-aligned self-similarity curves.}
For a user history $(i_1,\ldots,i_L)$ with item embeddings
$(e_{i_1},\ldots,e_{i_L})$, we compute the semantic self-similarity matrix
\[
    A_{pq} = \cos(e_{i_p}, e_{i_q}).
\]
For each lag $\Delta$, we average similarities along the corresponding diagonal,
\[
    s_u(\Delta) =
    \frac{1}{L-\Delta}
    \sum_{p=1}^{L-\Delta}
    \cos(e_{i_p}, e_{i_{p+\Delta}}).
\]
This produces a lag-aligned similarity curve for each user, where peaks at
small, medium, or large lags indicate recurring semantic structure at different
temporal ranges.

\paragraph{PCA over temporal modes.}
We collect lag-aligned similarity curves from long-history users and apply PCA
after centering the curves. The resulting principal components identify dominant
temporal modes in the similarity structure. To quantify how many temporal scales
are present for each user, we assign the peak lags of the leading modes to three
coarse lag bands: short, medium, and long. We then count the number of distinct
lag bands covered by the leading modes. On KuaiRec, 85\% of long-history users
have leading modes spanning at least two coarse lag bands, indicating that
multi-scale temporal structure is common rather than isolated to the example
shown in Figure~\ref{fig:motivation}.

\begin{figure}[H]
    \centering
    \includegraphics[width=0.98\linewidth]{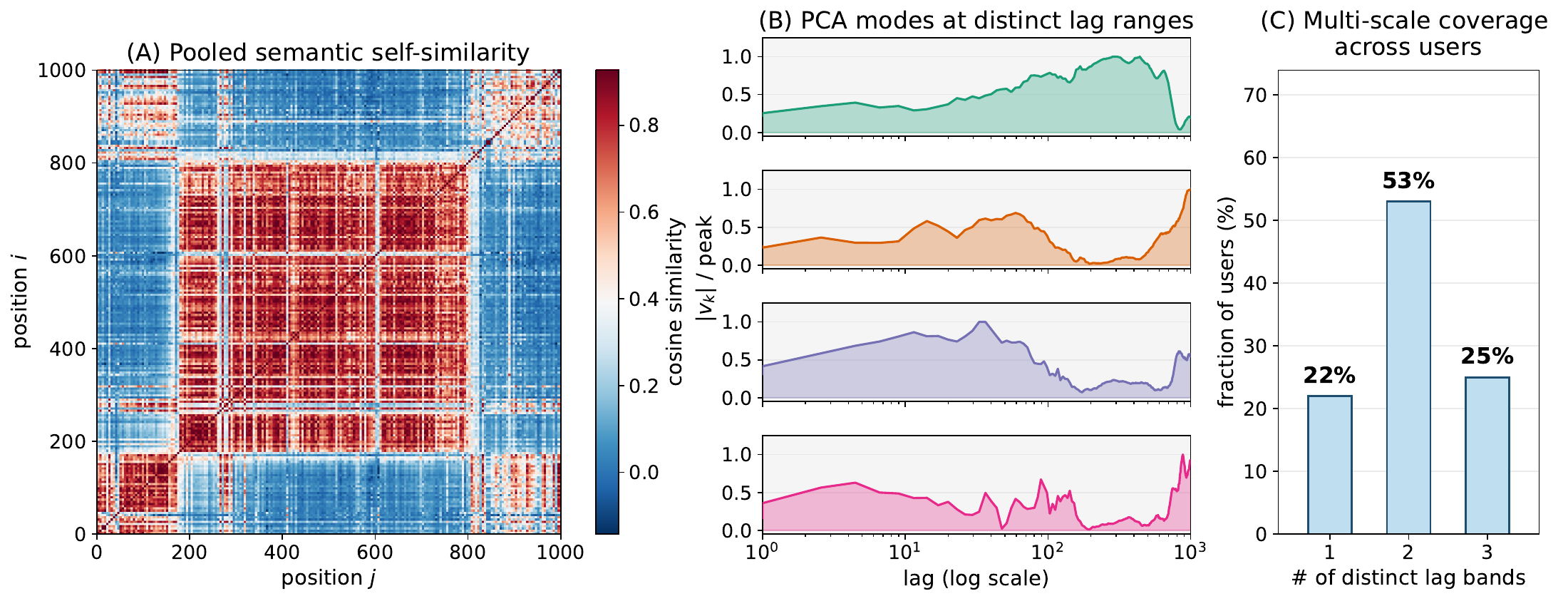}
    \caption{
\textbf{Analysis of multi-scale user-history structure.}
(A) Pooled semantic self-similarity for a long-history user shows coherent blocks at multiple temporal extents.
(B) PCA over lag-aligned similarity curves reveals dominant temporal loadings at distinct lag ranges. (C) Across long-history users, the peak lags of the top PCA modes typically cover multiple coarse temporal bands. These patterns motivate multi-resolution tracks and sparse seed-to-track routing.
}
    \label{fig:motivation2}
\end{figure}

\paragraph{KuaiRand-1K results.}
We repeat the same analysis on KuaiRand-1K. As shown in
Figure~\ref{fig:motivation2}, KuaiRand-1K exhibits the same qualitative pattern:
semantic self-similarity contains coherent blocks at multiple temporal extents,
and the leading lag-aligned modes cover multiple coarse lag ranges for most
long-history users. This supports the use of reusable user memory that organizes
behavioral evidence across temporal resolutions.

\newpage
\section{Industrial Dataset Result}
\label{app:industrial_validation}

We further evaluate MARS on an industrial dataset consisting of four anonymized tasks. We denote them as Task-A, Task-B, Task-C, and Task-D: Task-A corresponds to one consumption task, while Task-B, Task-C, and Task-D correspond to three engagement-event tasks. Due to confidentiality constraints, we anonymize task names and report only relative evaluation NE changes compared with VISTA.

All industrial comparisons use a chronological train/evaluation split: earlier logs are used for training, and a later contiguous time window is held out for offline evaluation.
MARS and VISTA use the same train/evaluation examples, item and user features, target-conditioned scorer and prediction head, upstream history preprocessing, and truncation or sampling setting.
Only the memory construction module is changed.
For MARS, we use $N_{\mathrm{mem}}=3$, $M=4$, $k=2$, $H_{\min}=5$, and $H_{\max}=2500$ under the maximum-history truncation of approximately $8$k interactions.
Due to the computational cost of industrial-scale training, we report one run for each industrial setting. These results therefore provide supplementary cross-task evidence but do not quantify seed variability; our main statistical conclusions rely on the public datasets, where results are averaged over three runs and accompanied by paired significance tests.

We evaluate two upstream history-processing settings. With history sampling disabled, the truncated history is consumed directly and the average effective history length is approximately $6$k. With history sampling enabled, older interactions are retained with lower probability and the average post-processed history length is approximately $1.5$k.

We report relative NE change compared with VISTA:
\[
\mathrm{Relative\ NE\ Change}
=
\frac{
\mathrm{NE}_{\mathrm{MARS}} - \mathrm{NE}_{\mathrm{VISTA}}
}{
\mathrm{NE}_{\mathrm{VISTA}}
}
\times 100\%.
\]
Lower NE is better, so negative values indicate improvement over VISTA.

\begin{table}[H]
\centering
\small
\caption{
Relative evaluation NE change compared with VISTA on industrial CTR prediction tasks (single run per setting).
Lower NE is better, negative values indicate improvement over VISTA.
}
\label{tab:industrial_relative_ne}
\setlength{\tabcolsep}{8pt}
\renewcommand{\arraystretch}{1.12}
\begin{tabular}{lcccc}
\toprule
Setting & Task-A & Task-B & Task-C & Task-D \\
\midrule
History sampling enabled  & $-1.50\%$ & $-3.10\%$ & $-19.67\%$ & $-16.16\%$ \\
History sampling disabled & $-1.96\%$ & $-5.00\%$ & $-18.08\%$ & $-18.52\%$ \\
\bottomrule
\end{tabular}
\end{table}

Across all four tasks and both history-processing settings, MARS reduces evaluation NE relative to VISTA. The gains remain consistent when history sampling is disabled and the effective history length increases from approximately $1.5$k to $6$k. In three of four tasks, the relative NE improvement is larger without history sampling. This trend is consistent with the public-benchmark analysis on KuaiRec, where MARS gains grow with effective history length. Overall, these results provide supplementary offline evidence beyond public datasets.

\newpage

\section{Additional Related Work}
\label{app:related_work}

\paragraph{Multi-scale state-space models.}
Multi-scale SSMs such as ms-Mamba~\citep{karadag2025msmamba} use multiple sampling rates, but target sequence backbones rather than candidate-independent cached user memory.

\paragraph{Memory-based long-history recommendation.}
MIMN~\citep{pi2019mimn} addresses lifelong user behavior modeling for CTR prediction through an industrial co-design of UIC and an incrementally updated multi-channel user-interest memory network.
User-memory networks~\citep{chen2018sequentialmemory} introduce explicit memory read/write modules for sequential recommendation.
MARS shares the goal of reusable memory, but factorizes memory by temporal retention scale and materializes compact seed memories through sparse seed-to-track routing.

\paragraph{Multi-interest user modeling.}
Several recommendation models represent a user with multiple interest vectors rather than a single embedding, but they differ in how the interest decomposition is used.
MIND~\citep{li2019mind} targets large-scale matching: it uses dynamic capsule routing to cluster historical behaviors into multiple interest capsules, and uses label-aware attention to learn target-relevant interest representations.
ComiRec~\citep{cen2020comirec} extends multi-interest extraction to sequential recommendation and candidate retrieval, with a controllable aggregation module that trades off accuracy and diversity.
DMIN~\citep{xiao2020dmin} targets CTR prediction: it refines behavior representations with self-attention and then extracts latent multiple interests for target-aware ranking.
The common abstraction in these methods is a set of semantic interest groups or intent slots.
This differs from cached summarization methods such as VISTA~\citep{chen2026massivememorizationhundredstrillions}, where multiple seed memories are used to cache long-history representations for candidate-independent scoring rather than primarily to expose semantic clusters.
MARS follows VISTA's cached-memory interface, but organizes the memory by temporal retention scale.
Each temporal track observes the full history with a different half-life prior, so seed specialization comes from selecting temporal resolutions rather than assigning behaviors to semantic interest groups.

\paragraph{Cached user-memory compression.}
Cached summarization methods decouple expensive history processing from candidate scoring.
VISTA~\citep{chen2026massivememorizationhundredstrillions} compresses lifelong histories into a fixed set of summary tokens that can be cached and reused by a downstream target-attention scorer.
Hi-SAM~\citep{pan2026hisam} also uses anchor tokens for cached compression, but its hierarchy is on the \emph{item-token side}, built around multi-modal semantic-ID structure, rather than on the \emph{temporal} side.
MARS adopts the same candidate-independent serving interface, but organizes reusable memory along temporal retention scales while leaving item tokenization unchanged.
Rather than producing monolithic seed memories, MARS first writes the full history into half-life-anchored recurrent temporal tracks and then materializes compact memory through sparse seed-to-track readout.
Outside recommendation, multimodal coverage maximization selects a compact subset of vision tokens to reduce the inference cost of vision-language models~\citep{dong2026mmtokmultimodalcoveragemaximization}; MARS instead compresses user histories into a reusable cached memory for candidate scoring.

\paragraph{LLM-based generative recommendation.}
Instruction-guided LLM generative recommenders must also decide how much to rely on interaction history, which can mislead generation when it conflicts with the user's current instruction~\citep{yin2026historymisleadsasymmetricmargin}.
MARS studies a complementary setting: CTR-style ranking with a candidate-independent cached memory, where the question is how to organize history across temporal scales before candidate scoring.

\paragraph{Sparse routing and mixture-of-experts.}
MARS uses sparse Top-$k$ routing with a straight-through Gumbel-Softmax estimator~\citep{jang2017categoricalreparameterizationgumbelsoftmax,dong2023weaklysupervisedvideorepresentation}.
This is related to sparse expert routing in MoE models~\citep{fedus2022switchtransformersscalingtrillion}, but the routed units play a different role.
MoE routing selects parameter experts to scale model capacity under limited per-token compute, often requiring load-balancing mechanisms.
MARS routes seed queries to temporal memory states that have already been constructed by the writer.
Thus, routing is not used to activate different parameter experts, but to materialize compact user memories while reducing dense mixing of short-, medium-, and long-range behavioral evidence.

\newpage

\section{Compressed State-Bank Variants}
\label{app:compressed_state_cache}

The default MARS configuration stores each temporal-track state as a full
$D\times D$ matrix.
We additionally evaluate compressed recurrent state variants to test
whether the optional update-state cache can be reduced without changing the
scorer-facing memory $\mathbf{H}_u$.
The diagonal variant stores a vector in $\mathbb{R}^{D}$ per track, while the
low-rank variants use projected $r\times r$ states.
These variants change only the optional state bank used for low-latency
incremental updates; candidate scoring still reads the same fixed-size
$\mathbf{H}_u$.

\begin{table}[H]
\centering
\small
\caption{
\textbf{Compressed recurrent state variants on KuaiRec.}
The update-state compression is measured relative to the default full
$D\times D$ per-track state. Results are from a single-run diagnostic, so we
treat them as supplementary serving evidence rather than replacing the default
full-state configuration.
}
\label{tab:compressed_state_bank}
\setlength{\tabcolsep}{5pt}
\renewcommand{\arraystretch}{1.08}
\begin{tabular}{lcccccc}
\toprule
Variant & Per-track state & Elements & Update-state comp. & Update cache & AUC $\uparrow$ & NE $\downarrow$ \\
\midrule
Full & $D\times D$ matrix & $1024$ & $1\times$ & $48$ KiB & $0.8041$ & $0.7913$ \\
Diagonal & $\mathbb{R}^{D}$ vector & $32$ & $32\times$ & $1.5$ KiB & $0.8012$ & $0.7999$ \\
$r=8$ & $r\times r$ matrix & $64$ & $16\times$ & $3$ KiB & $0.8012$ & $0.7996$ \\
$r=4$ & $r\times r$ matrix & $16$ & $64\times$ & $0.75$ KiB & $0.8003$ & $0.7984$ \\
$r=2$ & $r\times r$ matrix & $4$ & $256\times$ & $0.19$ KiB & $0.7940$ & $0.8028$ \\
\bottomrule
\end{tabular}
\end{table}

The diagonal, $r=8$, and $r=4$ variants retain most of the AUC gain over VISTA while substantially
reducing the optional update-state cache, whereas the $r=2$ variant loses most of that gain (AUC $0.7940$ vs.\ $0.7932$ for VISTA).
For example, with 4-byte FP32 values and the default
$N_{\mathrm{mem}}=3,M=4,D=32$ setting, the full update-state bank is $48$ KiB
per user, while the $r=4$ variant reduces it to $0.75$ KiB.
The total model-side cache still includes the scorer-facing $16$ KiB
$\mathbf{H}_u$, so the corresponding total cache changes from $64$ KiB to
$16.75$ KiB rather than receiving the full $64\times$ reduction.
Since these results are not averaged over random seeds, we report them as an
optional deployment trade-off and keep the full state bank as the default
configuration in the main experiments.

\newpage
\section{Parallel Scan Derivation}
\label{app:parallel_scan}

Direct sequential evaluation of Eq.~\eqref{eq:state_update} requires
$O(L)$ sequential steps per track and per layer, which becomes a bottleneck
for training on long histories. To enable efficient training, we adopt the
chunk-wise parallel scan formulation of gated linear attention~\citep{yang2024gatedlinearattentiontransformers},
which we derive here for completeness.

For clarity, consider a single track and omit the track and layer indices. The
recurrence is
\[
    \mathbf{S}_t =
    (\mathbf{g}_t \mathbf{1}^{\top}) \odot \mathbf{S}_{t-1}
    + \mathbf{k}_t \mathbf{v}_t^{\top},
\]
with $\mathbf{S}_t \in \mathbb{R}^{D\times D}$,
$\mathbf{g}_t, \mathbf{k}_t, \mathbf{v}_t \in \mathbb{R}^{D}$, and
$\mathbf{1} \in \mathbb{R}^{D}$. Since the gate $\mathbf{g}_t\mathbf{1}^{\top}$
is broadcast across columns, the recurrence decouples across the $D$ row
dimensions:
\[
    \mathbf{S}_{t,r,:}
    =
    g_{t,r}\,\mathbf{S}_{t-1,r,:}
    + k_{t,r}\,\mathbf{v}_t^{\top},
    \qquad r = 1,\dots,D.
\]
Unrolling gives
\[
    \mathbf{S}_{t,r,:}
    =
    \sum_{m=1}^{t}
    \left(
    \prod_{n=m+1}^{t} g_{n,r}
    \right)
    k_{m,r}\,\mathbf{v}_m^{\top},
\]
with the convention that the empty product equals $1$. The product term can be
computed from prefix sums in log-space:
\[
    \prod_{n=m+1}^{t} g_{n,r}
    =
    \exp\!\left(G_{t,r} - G_{m,r}\right),
    \qquad
    G_{t,r}=\sum_{n=1}^{t}\log g_{n,r}.
\]
Hence the decay weights within a chunk can be obtained from prefix differences
rather than sequential multiplication.

\paragraph{Chunk-wise composition.}
We partition the sequence into chunks of length $C$. Within each chunk, decay
weights are computed from the log-prefix values above. Across chunks, the
recurrence reduces to an affine transition
\[
    \mathbf{S}_{b}^{\mathrm{out}}
    =
    \mathbf{A}_b \odot \mathbf{S}_{b}^{\mathrm{in}}
    + \mathbf{B}_b,
\]
where $\mathbf{A}_b$ is the cumulative gate product over chunk $b$, broadcast
across columns, and $\mathbf{B}_b$ is the locally accumulated update. Two
adjacent chunks compose associatively:
\[
    (\mathbf{A}_2,\mathbf{B}_2)\circ(\mathbf{A}_1,\mathbf{B}_1)
    =
    \bigl(
    \mathbf{A}_2\odot \mathbf{A}_1,\,
    \mathbf{A}_2\odot \mathbf{B}_1+\mathbf{B}_2
    \bigr).
\]
This associative operator enables a standard parallel prefix scan over chunks,
while within-chunk states use the log-prefix formulation above. The total work
is linear in the sequence length $L$ for fixed $D$, $M$, and number of memory
layers, and the chunk-level scan has parallel depth $O(\log(L/C))$. In all
experiments, we use $C=64$.

\newpage
\section{Realized Half-Life Analysis}
\label{app:realized_halflife}

The half-life anchors in \Eqref{eq:halflife_bias} fix only the \emph{base} retention scale of each track, whereas the realized gate in \Eqref{eq:gate} also contains an unconstrained input-dependent offset $\mathbf{F}_g(\mathbf{x}_t)$.
We therefore verify that training does not collapse the tracks toward a common timescale or exchange their temporal order.

\paragraph{Protocol.}
We run the trained KuaiRec model on $1{,}280$ held-out users and record the realized forget gates $g^{(i,\ell)}_{t,d}$ for every track $i$, memory layer $\ell$, valid token $t$, and state dimension $d$.
Each gate value is converted into an instantaneous half-life, measured in steps,
\[
H^{(i,\ell)}_{t,d}=\frac{\log(1/2)}{\log g^{(i,\ell)}_{t,d}},
\]
which equals the anchor $H_i$ when the input-dependent offset is zero.

\paragraph{Realized ranges.}
Table~\ref{tab:temporal_org}(a) in the main text reports, for each track $i$, the range $[L_i,U_i]$ from the minimum p05 to the maximum p95 of $H^{(i,\ell)}_{t,d}$ across the three memory layers, together with the maximum relative endpoint deviation $\max(|L_i-H_i|,|U_i-H_i|)/H_i$.
All endpoints lie within $8.1\%$ of their anchors, and the four ranges are disjoint, with adjacent bands separated by a factor of at least about $4.5$.
The adaptation is most visible in the slowest track at the scorer-facing layer, whose p05--p95 half-life range is $574.1$--$650.5$ steps while remaining clearly separated from the other tracks.

\paragraph{Pointwise ordering.}
Beyond these marginal ranges, we check the ordering exhaustively over $71{,}031{,}712$ valid token--dimension positions per layer, or $213{,}095{,}136$ positions in total.
Across all three memory layers, we observe zero violations of
\[
H^{(1,\ell)}_{t,d}<H^{(2,\ell)}_{t,d}<H^{(3,\ell)}_{t,d}<H^{(4,\ell)}_{t,d}.
\]

\paragraph{Why the ordering is retained.}
In pre-sigmoid logit space, the frozen anchor biases are approximately $1.91$, $3.58$, $5.19$, and $6.81$, with a minimum adjacent separation of $1.61$.
The learned offset is small in comparison: its median norm is below $0.3\%$ of the anchor-bias norm, and its p05--p95 spread is at most $0.15$ per track.
Because the sigmoid and the map from gate value to half-life are both strictly increasing, ordering in logit, gate, and half-life space is equivalent, and offsets of this size are far too small to exchange adjacent tracks, consistent with the zero violations above.
The input-dependent term therefore provides local adaptation within each temporal band without collapsing the temporal basis. Its contribution to accuracy is quantified by the constant-anchor control in Section~\ref{subsec:ablation} and Appendix~\ref{app:component_details}.

\newpage
\section{Routing Stability Diagnostics}
\label{app:routing_diagnostics}

We further diagnose whether sparse seed-to-track routing collapses to a small
subset of temporal tracks or remains adaptive across users.
We run MARS once over the full KuaiRec test set in evaluation mode and record
the deterministic hard Top-$k$ routing masks
$h_{u,j,i}^{(\ell)}\in\{0,1\}$ for user $u$, seed $j$, track $i$, and memory
layer $\ell$. Here $h_{u,j,i}^{(\ell)}=1$ indicates that track $i$ is selected
by seed $j$ for user $u$ at layer $\ell$.

For each layer, we compute the marginal track selection rate
\[
p_i^{(\ell)}
=
\frac{1}{NJ}
\sum_{u=1}^{N}
\sum_{j=1}^{J}
h_{u,j,i}^{(\ell)},
\qquad
\tilde p_i^{(\ell)}=\frac{p_i^{(\ell)}}{k},
\]
where $N$ is the number of users and each seed selects $k$ tracks, so
$\sum_i p_i^{(\ell)}=k$ and $\sum_i\tilde p_i^{(\ell)}=1$.
We report the entropy
\[
H^{(\ell)}
=
-\sum_{i=1}^{M}
\tilde p_i^{(\ell)}\log \tilde p_i^{(\ell)},
\]
the effective number of tracks $\exp(H^{(\ell)})$, and the cross-user routing
variance
\[
V^{(\ell)}
=
\frac{1}{JM}
\sum_{j=1}^{J}
\sum_{i=1}^{M}
\mathrm{Var}_{u}
\!\left[
h_{u,j,i}^{(\ell)}
\right].
\]

\begin{table}[H]
\centering
\small
\caption{
\textbf{Routing stability diagnostics on KuaiRec.}
We report final-layer track usage because it directly produces the scorer-facing cached memory $\mathbf{H}$, and report cross-user routing variance $V^{(\ell)}$ across layers to show how routing becomes history-conditioned.
Normalized track rates are $\tilde p_i^{(\ell)}=p_i^{(\ell)}/k$, where $p_i^{(\ell)}$ is the fraction of seeds whose hard Top-$k$ routing selects track $i$ at layer $\ell$.
}
\label{tab:routing_diagnostics}
\setlength{\tabcolsep}{8pt}
\renewcommand{\arraystretch}{1.12}

\begin{minipage}[t]{0.62\linewidth}
\centering
\vspace{0.35em}
\textbf{(a) Final-layer track usage} ($M{=}4$, $k{=}2$)
\vspace{0.35em}

\begin{tabular}{ccc}
\toprule
Entropy & Effective tracks & Normalized track rates $\tilde p^{(L_{\mathrm{mem}})}$ \\
\midrule
$1.366$ & $3.92$ & $[0.20,\ 0.20,\ 0.31,\ 0.29]$ \\
\bottomrule
\end{tabular}
\end{minipage}
\hfill
\begin{minipage}[t]{0.32\linewidth}
\centering
\textbf{(b) Cross-user routing variance}
\vspace{0.35em}

\begin{tabular}{cc}
\toprule
Layer & $V^{(\ell)}$ \\
\midrule
1 & $0.0000$ \\
2 & $0.0063$ \\
3 & $0.1670$ \\
\bottomrule
\end{tabular}
\end{minipage}
\end{table}

The final layer uses all four temporal tracks in a balanced manner (effective tracks $3.92/4$, normalized rates between $0.20$ and $0.31$), confirming the absence of track collapse in the cached memory. Cross-user routing variance grows monotonically across layers: it is exactly zero at layer 1, where seeds are input-independent by construction, and reaches $0.167$ at the final layer, where seeds have received history-specific information through the residual updates of Eq.~\ref{eq:seed_update}. This confirms the hierarchical history-conditioning claim of Section~\ref{subsec:routing}.

The first-layer routing is user-independent by construction, since the initial
seed queries are learned parameters before any history-conditioned readout;
accordingly, $V^{(1)}=0$.
After each routed readout, residual seed updates inject user-specific history
information, and the routing variance increases in deeper layers.
At the final layer, the effective number of selected tracks is $3.92$ out of the
maximum $4$, and all normalized track rates lie between $0.20$ and $0.31$.
This indicates that the router does not exhibit marginal track collapse on the
test set, while deeper routing becomes increasingly history-conditioned.

\newpage
\section{Routing Participation versus History Length}
\label{app:routing_length}

To further examine whether learned routing can be explained by history length
alone, we analyze the relationship between per-track participation and history
length on KuaiRec.

For each user $u$ and temporal track $i$, we define track participation as
\[
p_{u,i}
=
\frac{1}{J}
\sum_{j=1}^{J}
\mathbf{1}\left[i \in \mathrm{TopK}(z_{u,j})\right],
\]
where $J$ is the number of virtual seeds and $z_{u,j}$ denotes the routing logits
for seed $j$. Thus, $p_{u,i}$ measures the fraction of seeds whose hard Top-$k$
routing selects track $i$. Since each seed selects $k$ tracks, participation
values across tracks are not normalized to sum to one.

\begin{figure}[!htbp]
    \centering
    \includegraphics[width=0.75\textwidth]{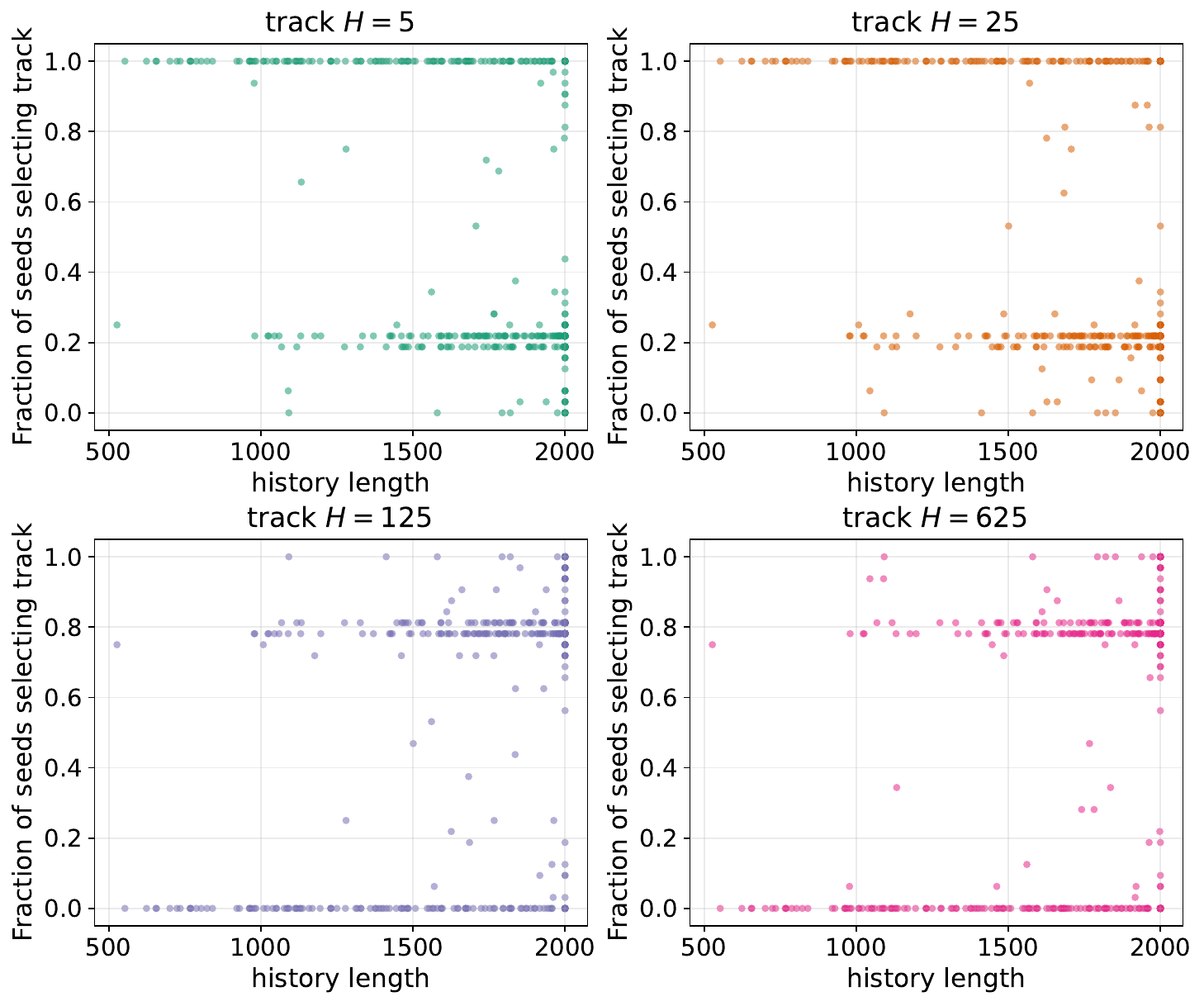}
    \caption{
    Per-track routing participation versus history length for 512 KuaiRec users with
    history lengths between 500 and 2{,}000. Each point corresponds to one user. The $y$-axis
    shows the fraction of virtual seeds whose hard Top-$k$ routing selects the
    corresponding temporal track. Although history length is related to track
    usage, the learned routing does not collapse to a simple length-based rule:
    short-half-life tracks remain active for many long-history users,
    long-half-life tracks are not uniformly selected at large history lengths,
    and users with similar history lengths can exhibit different
    track-participation patterns.
    }
    \label{fig:routing_length}
\end{figure}

We analyze 512 KuaiRec users with history lengths between 500 and 2{,}000, which covers the main
long-history regime in KuaiRec and is consistent with the history-length
partition used in the main analysis.
Figure~\ref{fig:routing_length} shows that
the learned routing does not reduce to a simple length-based shortcut.
Short-half-life tracks remain active for many long-history users, while
long-half-life tracks are not uniformly selected at large history lengths.
Moreover, users with similar history lengths can exhibit different
track-participation patterns. This supports the interpretation that routing
reflects user-specific behavioral structure beyond history length alone.

\newpage
\section{Routing and Robustness Under Behavioral Shift}
\label{app:shift}

\paragraph{Routing response to recent behavior.}
For $4{,}096$ held-out long-history users, we compute the final-layer hard routing mask from the full history and from the history with the most recent $30$ interactions removed, using the same trained model at inference time.
Users are divided into drift terciles, with about $1{,}365$ users per tercile, according to the cosine distance between the mean item embedding of the recent segment and that of the preceding history.
The overall before/after routing-mask Jaccard similarity is $0.934$.
Table~\ref{tab:shift_routing} compares the lowest and highest drift terciles, where the short-track share is the fraction of routing selections assigned to the $H=5$ and $H=25$ tracks.

\begin{table}[H]
\centering
\small
\caption{\textbf{Routing response to recent behavior.} Differences use a user-level bootstrap; both differences have $p<0.001$.}
\label{tab:shift_routing}
\setlength{\tabcolsep}{5pt}
\begin{tabular}{lcccc}
\toprule
Metric & Low drift & High drift & High $-$ low & 95\% CI \\
\midrule
Before/after mask Jaccard & $0.942$ & $0.931$ & $-0.011$ & $[-0.016, -0.006]$ \\
Increase in short-track share & $+0.047$ & $+0.055$ & $+0.008$ & $[+0.003, +0.012]$ \\
\bottomrule
\end{tabular}
\end{table}

Routing thus remains stable when recent behavior is consistent with the preceding history, and reallocates capacity toward shorter temporal resolutions when recent behavior departs more strongly from earlier interests.

\paragraph{Shift-point selection.}
For each of $4{,}000$ users, we select one shift point using a model-independent drift score computed from adjacent interaction windows. The procedure does not use model predictions, losses, evaluation metrics, or routing weights.
We evaluate the same $20$ interactions before and after each shift point, yielding $80{,}000$ pre-shift and $80{,}000$ post-shift examples.
All models use their trained checkpoints without retraining or fine-tuning.

\paragraph{Paired advantage.}
Let $A(t)$ denote the MARS-minus-baseline metric difference in period $t\in\{\mathrm{pre},\mathrm{post}\}$, and let $G=A(\mathrm{post})-A(\mathrm{pre})$.
Positive $G_{\mathrm{AUC}}$ and negative $G_{\mathrm{NE}}$ favor MARS.
Confidence intervals and $p$-values use $10{,}000$ paired user-level bootstrap resamples.
The user-independent Top-$k$ control is trained independently and preserves the multi-resolution tracks, capacity, routing budget, readout, scorer, and training protocol, so its comparison with MARS isolates history-conditioned routing.
Table~\ref{tab:shift_full} gives the full results.

\begin{table}[H]
\centering
\caption{\textbf{Paired advantage of MARS before and after behavioral shifts}, with 95\% paired user-level bootstrap CIs. All gaps have $p<0.001$.}
\label{tab:shift_full}
\resizebox{\linewidth}{!}{%
\begin{tabular}{llccc}
\toprule
Baseline & Metric & Pre-shift $A$ & Post-shift $A$ & Robustness gap $G$ \\
\midrule
VISTA & AUC $\uparrow$ & $+0.006$ $[+0.002, +0.010]$ & $+0.017$ $[+0.011, +0.023]$ & $+0.011$ $[+0.006, +0.016]$ \\
 & NE $\downarrow$ & $-0.008$ $[-0.014, -0.002]$ & $-0.027$ $[-0.036, -0.018]$ & $-0.019$ $[-0.028, -0.010]$ \\
\midrule
User-independent Top-$k$ & AUC $\uparrow$ & $+0.0004$ $[-0.0002, +0.0010]$ & $+0.0036$ $[+0.0025, +0.0047]$ & $+0.0032$ $[+0.0020, +0.0044]$ \\
 & NE $\downarrow$ & $-0.0008$ $[-0.0018, +0.0002]$ & $-0.0056$ $[-0.0074, -0.0038]$ & $-0.0048$ $[-0.0065, -0.0031]$ \\
\bottomrule
\end{tabular}}
\end{table}

MARS therefore has a significantly larger advantage over VISTA after the selected behavioral shifts.
Against the user-independent control, the routing benefit is small and statistically inconclusive before the shift but significant after it, which attributes part of the post-shift robustness gain specifically to history-conditioned routing.
We scope this conclusion to the evaluated within-user behavioral shifts rather than arbitrary out-of-distribution settings.

\newpage
\section{Component Decomposition, Additional Baselines, and Recoverability Tests}
\label{app:component_details}

\paragraph{Controls.}
All controls in Table~\ref{tab:ablation_summary}(a) use the same target-conditioned scorer, prediction head, and training protocol as MARS on KuaiRec.
With $M=1$, routing is degenerate because only one track is available, so a ``single trajectory with routing'' control reduces to the single-track row.
The \emph{single-resolution bank} therefore provides the nontrivial capacity-matched routing control: its $M=4$ tracks share the same retention scale but keep independent content projections, which preserves track count, capacity, projections, routing, readout, scorer, and training pipeline while removing temporal-resolution diversity.
The multi-resolution bank with uniform readout is the ``multiple trajectories without routing'' control, and user-independent versus user-conditioned Top-$k$ isolates history-conditioned adaptation.

\paragraph{Paired bootstrap tests.}
Using identical held-out examples, we apply paired user-level bootstrap tests with $\Delta\mathrm{AUC}=\mathrm{AUC}_{\mathrm{MARS}}-\mathrm{AUC}_{\mathrm{control}}$ and $\Delta\mathrm{NE}=\mathrm{NE}_{\mathrm{MARS}}-\mathrm{NE}_{\mathrm{control}}$ (Table~\ref{tab:bootstrap_ablation}).
All confidence intervals exclude zero in the direction favoring MARS.
The constant-anchor control removes the input-dependent gate offset, $\mathbf{g}_t=\sigma(\mathbf{b}_g)$, while preserving the rest of the architecture and training protocol.
These tests quantify evaluation-set uncertainty rather than training-seed variability.

\begin{table}[H]
\centering
\small
\caption{\textbf{Paired user-level bootstrap tests for the ablations on KuaiRec.} $\Delta=\mathrm{MARS}-\mathrm{control}$; positive $\Delta$AUC and negative $\Delta$NE favor MARS. All 95\% CIs exclude zero.}
\label{tab:bootstrap_ablation}
\setlength{\tabcolsep}{3pt}
\begin{tabular}{lcc}
\toprule
Control & $\Delta$AUC [95\% CI] & $\Delta$NE [95\% CI] \\
\midrule
Single-res.\ bank ($M{=}4$), Top-$k$ & $+0.007$ $[+0.004, +0.010]$ & $-0.013$ $[-0.018, -0.008]$ \\
Multi-res.\ bank ($M{=}4$), uniform & $+0.007$ $[+0.004, +0.010]$ & $-0.015$ $[-0.020, -0.010]$ \\
Single track ($M=1$) & $+0.010$ $[+0.007, +0.013]$ & $-0.021$ $[-0.027, -0.015]$ \\
Constant-anchor gates & $+0.0009$ $[+0.0003, +0.0015]$ & $-0.0022$ $[-0.0034, -0.0010]$ \\
\bottomrule
\end{tabular}
\end{table}

\paragraph{Concatenated pooled-window baseline.}
This baseline constructs masked mean-pooled summaries over suffix windows of $\{30, 200, 1000, 2000\}$ interactions, concatenates them into a candidate-independent cached memory, and uses the same scorer and prediction head as MARS. It is trained on KuaiRec with the same recipe and three random seeds (Table~\ref{tab:pooled_windows}).
The baseline improves over VISTA, which supports the temporal-aliasing diagnosis: explicitly retaining several temporal summaries recovers information lost under monolithic compression.
It nevertheless remains below MARS. Pooled windows commit to a few hard, index-based cutoffs during memory construction, so interactions with similar recency can be separated simply because they fall on opposite sides of a boundary, whereas MARS applies continuous half-life retention over the full sequence and lets each seed select the relevant resolutions during readout.

\begin{table}[H]
\centering
\small
\caption{\textbf{Concatenated pooled-window baseline on KuaiRec} (mean $\pm$ std over three seeds).}
\label{tab:pooled_windows}
\begin{tabular}{lcc}
\toprule
Method & AUC $\uparrow$ & Gain over VISTA \\
\midrule
VISTA & $0.793\pm0.002$ & -- \\
Concatenated pooled windows & $0.797\pm0.001$ & $+0.004$ \\
MARS & $\mathbf{0.804\pm0.002}$ & $\mathbf{+0.011}$ \\
\bottomrule
\end{tabular}
\end{table}

\paragraph{Fixed-window recoverability.}
We extend the probing protocol of Section~\ref{sec:temporal_probe} to the fixed-window memory of Figure~\ref{fig:analysis_all}(b), using the same $512$ long-history KuaiRec users, PPMI-SVD item representations, 5-fold Ridge probes, and band definitions (Table~\ref{tab:temporal_org}(b) in the main text).
The users are selected by the predefined long-history criterion rather than by model outcomes.
Fixed windows reduce the long-range bias of VISTA, but retain less short- and medium-range information than MARS and have a lower mean recoverability.
Every MARS track observes the full sequence through a continuous retention kernel, providing overlapping soft resolutions rather than discrete window boundaries.
A paired user-level bootstrap over the same users, using out-of-fold probe predictions, gives a mean-$R^2$ difference (MARS minus fixed-window) of $+0.022$ with 95\% CI $[+0.010, +0.034]$ ($p<0.05$), and a cross-band-range reduction (fixed-window range minus MARS range) of $+0.090$ with 95\% CI $[+0.050, +0.130]$ ($p<0.05$).
The probe explains the downstream results rather than replacing recommendation evaluation.

\newpage

\section{Additional Hyperparameter Sensitivity}
\label{app:hyper_sensitivity}

We further evaluate the sensitivity of MARS to the seed capacity $J$ and the number of memory layers $N_{\mathrm{mem}}$ on KuaiRec.

All other hyperparameters are fixed to the default setting when varying one factor.

\begin{figure}[H]
    \centering
    \includegraphics[width=\linewidth]{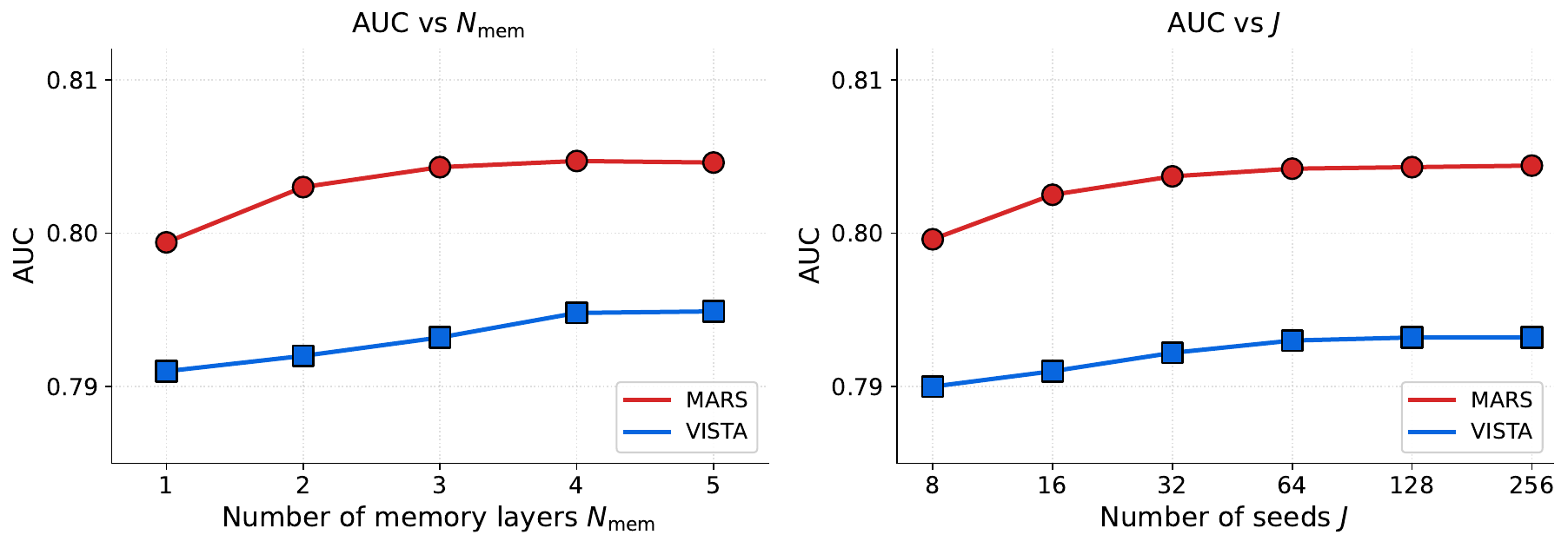}
    \caption{
    \textbf{Sensitivity to memory depth and seed capacity on KuaiRec.}
    We vary the number of memory layers $N_{\mathrm{mem}}$ and the number of virtual seeds $J$ while keeping other hyperparameters fixed.
    Both MARS and VISTA improve with larger capacity and then plateau, but MARS consistently outperforms VISTA across all settings.
    }
    \label{fig:j_nmem_sensitivity}
\end{figure}

Figure~\ref{fig:j_nmem_sensitivity} shows that increasing $N_{\mathrm{mem}}$ improves performance up to a small number of memory layers and then saturates.
Similarly, increasing $J$ improves AUC up to moderate seed capacity, with diminishing returns beyond $J=64$.
MARS consistently outperforms VISTA across both sweeps, indicating that the improvement is not explained by simply increasing memory depth or scorer-facing seed capacity.

\begin{table}[H]
\centering
\small
\caption{
Hyperparameter settings used in the experiments.
}
\label{tab:hyperparams}
\begin{tabular}{lccc}
\toprule
\textbf{Hyperparameter} & \textbf{Amazon} & \textbf{KuaiRand} & \textbf{KuaiRec} \\
\midrule
\multicolumn{4}{l}{\textit{Shared / VISTA-aligned settings}} \\
Learning rate & $5.0{\times}10^{-4}$ & $1.0{\times}10^{-4}$ & $5.0{\times}10^{-4}$ \\
Optimizer & Adam & Adam & Adam \\
Batch size & 1024 & 1024 & 1024 \\
Batch norm & No & No & No \\
Early-stop patience & 4 & 5 & 5 \\
Embedding regularizer & 0.005 & None & None \\
Embedding dimension $D$ & 64 & 32 & 32 \\
Embedding initializer & $1.0{\times}10^{-4}$ & $1.0{\times}10^{-4}$ & $1.0{\times}10^{-4}$ \\
MLP hidden units & [1024, 512, 256] & [512, 128, 64] & [512, 128, 64] \\
MLP activation & ReLU & ReLU & ReLU \\
Number of seeds $J$ & 128 & 128 & 128 \\
Memory layers & 1 & 3 & 3 \\
Scorer attention heads & 4 & 4 & 4 \\
Scorer attention layers & 1 & 1 & 1 \\
\midrule
\multicolumn{4}{l}{\textit{MARS-specific settings}} \\
Temporal tracks $M$ & 4 & 4 & 4 \\
Routing budget $k$ & 2 & 2 & 2 \\
Half-life anchors $\{H_i\}$ & $\{2,6,21,64\}$ & $\{5,25,125,625\}$ & $\{5,25,125,625\}$ \\
Router type & ST Top-$k$ & ST Top-$k$ & ST Top-$k$ \\
Training-time routing noise & Gumbel & Gumbel & Gumbel \\
Inference routing & Det. Top-$k$ & Det. Top-$k$ & Det. Top-$k$ \\
\bottomrule
\end{tabular}
\end{table}

\newpage
\section{Cache Footprint Details}
\label{app:cache_footprint}

This appendix provides the byte-level cache-footprint details for the serving modes discussed in Section~\ref{sec:efficiency}.
We count user-specific model-side cached states only, excluding shared model parameters, item embeddings, storage-system metadata, and feature-store overhead.
The goal is to quantify the per-user state that must be stored to support candidate scoring and, optionally, low-latency incremental user-state updates.

Both VISTA and MARS use a cached compact user memory for candidate scoring.
In both cases, the target-conditioned scorer reads a fixed-size memory $\mathbf{H}_u \in \mathbb{R}^{J \times D}$.
Therefore, the scorer-facing cache contains $JD$ values per user and is independent of the raw history length $L$.
This is the cache used on the candidate-scoring path.

MARS can optionally persist an additional recurrent state bank for low-latency incremental updates,
\[
\mathcal{S}_u = \{ \mathbf{S}^{(i,\ell)}_{u,L} \}_{i=1,\ell=1}^{M,N_{\mathrm{mem}}},
\]
where each temporal-track state has shape $D \times D$.
With $M$ temporal tracks and $N_{\mathrm{mem}}$ memory layers, this optional update cache contains $N_{\mathrm{mem}}MD^2$ values per user.
If each value uses $b$ bytes, the per-user cache footprints are
\[
\mathrm{Cache}_{\mathrm{score}} = bJD,
\qquad
\mathrm{Cache}_{\mathrm{incremental}} = b(JD + N_{\mathrm{mem}}MD^2).
\]
The state bank $\mathcal{S}_u$ is not read by the candidate scorer.
It is only used to update the user memory after a new interaction without replaying the full history.
Thus, persisting $\mathcal{S}_u$ is a storage--latency trade-off on the update path, not a change to the candidate-scoring interface.

\begin{table}[!htbp]
\centering
\small
\caption{
\textbf{Per-user model-side cache footprint under the default setting.}
We use $J=128$, $D=32$, $M=4$, $N_{\mathrm{mem}}=3$, and assume 4-byte FP32 values. 
Both VISTA and MARS cache the scorer-facing compact memory $\mathbf{H}_u$.
MARS can optionally cache the recurrent state bank $\mathcal{S}_u$ for low-latency incremental updates.
}
\label{tab:cache_footprint_bytes}
\begin{tabular}{lccc}
\toprule
Model / mode & Scorer cache $\mathbf{H}_u$ & Update cache $\mathcal{S}_u$ & Total \\
\midrule
VISTA summary-cache mode & $16$ KiB & -- & $16$ KiB \\
MARS refresh mode & $16$ KiB & -- & $16$ KiB \\
MARS incremental-update mode & $16$ KiB & $48$ KiB & $64$ KiB \\
\bottomrule
\end{tabular}
\end{table}

Under the default setting, MARS refresh mode has the same scorer-facing cache footprint as VISTA.
MARS incremental-update mode stores an additional $48$ KiB recurrent state bank per user, increasing the total model-side cache to $64$ KiB per user.
In memory-constrained deployments, this optional state bank can be persisted only for recently active users, quantized, evicted with a time-to-live policy, or recomputed asynchronously during periodic refresh.
These choices affect update freshness and update latency, while candidate scoring remains unchanged because it only reads $\mathbf{H}_u$.

\paragraph{Choosing a serving mode.}
Table~\ref{tab:serving_modes} summarizes the storage--accuracy options on KuaiRec.
Periodic refresh suffices when immediate event-level updates are unnecessary; low-rank ($r=4$) or diagonal states suit incremental updates under memory constraints; and full states give the best accuracy with immediate updates.
The $r=4$ state compresses the optional update state by $64\times$ and adds only $0.75$ KiB per user over VISTA while retaining most of the gain.
For scale, maintaining incremental states for $100$ million active users would require about $4.5$ TiB for the full optional state before replication, versus about $72$ GiB for the $r=4$ state.
The recurrent state is therefore best treated as a configurable hot-user cache rather than a mandatory full-population cache, and the $64$ KiB total is a storage--freshness--accuracy choice rather than a fixed $4\times$ increase.

\begin{table}[H]
\centering
\small
\caption{\textbf{Storage--accuracy options for serving on KuaiRec} (FP32, default setting). Compressed-state results are single-run (Appendix~\ref{app:compressed_state_cache}).}
\label{tab:serving_modes}
\begin{tabular}{lcc}
\toprule
Serving mode & Total cache / user & AUC $\uparrow$ \\
\midrule
VISTA & $16$ KiB & $0.7932$ \\
MARS, periodic refresh & $16$ KiB & $0.8041$ \\
MARS, incremental, low-rank $r=4$ & $16.75$ KiB & $0.8003$ \\
MARS, incremental, diagonal & $17.5$ KiB & $0.8012$ \\
MARS, incremental, full state & $64$ KiB & $0.8041$ \\
\bottomrule
\end{tabular}
\end{table}

\newpage
\section{Measured Serving Latency}
\label{app:latency}

\paragraph{Setup.}
We measure wall-clock latency on one NVIDIA A100 with batch size $256$, $D=32$, $J=128$, $N_{\mathrm{mem}}=3$, $M=4$, and $k=2$.
History initialization is performed beforehand and excluded from all timed regions, so the benchmark characterizes warm-cache serving.
The history-length experiment times one recurrent-state update and seed-memory materialization; end-to-end latency additionally includes scoring $N$ candidates per user.
We report medians over $40$ runs, and p95 ratios for end-to-end latency; the coefficient of variation is $0.7\%$--$1.4\%$.

\paragraph{Cost decomposition.}
A warm-cache request with $N$ candidates decomposes as $T(N)=T_{\mathrm{update}}+N\,T_{\mathrm{score}}$.
For MARS, one recurrent update costs $O(N_{\mathrm{mem}}MD^2)$ and seed-memory materialization costs $O(N_{\mathrm{mem}}JkD^2)$; both are independent of $N$ and of the cached history length.
VISTA and MARS expose the same $J\times D$ memory to the same target-conditioned scorer, whose dominant candidate-dependent cost is $O(JD)$ per candidate and $O(NJD)$ for $N$ candidates.
Both methods therefore scale linearly in $N$, and the fixed update-cost difference is amortized as the candidate pool grows.
Across the tested pool sizes, the measured MARS/VISTA scoring-only latency ratio remains within $1.009$--$1.014\times$, consistent with the shared scorer computation.

\begin{table}[H]
\centering
\small
\caption{\textbf{MARS update-and-materialization latency versus history length} (ms per batch of $256$ users, median over $40$ runs).}
\label{tab:latency_history}
\begin{tabular}{lccccc}
\toprule
History length $S$ & $500$ & $1{,}000$ & $2{,}000$ & $3{,}000$ & $5{,}000$ \\
\midrule
Latency (ms) & $0.421$ & $0.422$ & $0.423$ & $0.423$ & $0.425$ \\
\bottomrule
\end{tabular}
\end{table}

Across a $10\times$ increase in history length, update-and-materialization latency changes by less than $1\%$, confirming that warm-cache serving depends on fixed-size states rather than the raw history length.

\begin{table}[H]
\centering
\small
\caption{\textbf{End-to-end warm-cache latency} (batch of $256$ users) versus the number of candidates $N$ per user.}
\label{tab:latency_e2e}
\begin{tabular}{lcccc}
\toprule
Candidates $N$ & MARS median & VISTA median & Median ratio & p95 ratio \\
\midrule
$1$ & $0.588$ ms & $0.469$ ms & $1.253\times$ & $1.260\times$ \\
$100$ & $2.043$ ms & $1.906$ ms & $1.072\times$ & $1.075\times$ \\
$1{,}000$ & $15.478$ ms & $15.152$ ms & $1.021\times$ & $1.022\times$ \\
\bottomrule
\end{tabular}
\end{table}

Large-scale recommenders typically narrow the candidate pool through a multi-stage cascade, from millions or billions of items to tens of thousands, thousands, and eventually hundreds or fewer candidates across retrieval, pre-ranking, ranking, and re-ranking~\citep{li2024contextual,song2022rethinking}.
$N=1{,}000$ therefore reflects a practical ranking-stage workload.
At this setting, MARS requires about $1.02\times$ the end-to-end latency of VISTA at both the median and p95, and processes about $16.5$ million candidate evaluations per second on one A100.

\paragraph{Scope.}
These measurements support warm-cache serving efficiency for the evaluated A100 implementation, with candidate scoring independent of the raw history length.
They exclude one-time initialization from the complete raw history; initialization, multi-platform benchmarking, and full production-stack evaluation remain future work.

\newpage

\section{Serving Interfaces of Compared Methods}
\label{app:serving_interfaces}
\begin{table}[H]
\centering
\small
\caption{
\textbf{Serving-interface comparison.}
Only cached user-memory methods expose the same candidate-independent scorer-facing memory $\mathbf{H}_u$, so Section~\ref{sec:efficiency} reports quantitative cache and update-cost comparisons for this family.
}
\label{tab:serving_interface_comparison}
\setlength{\tabcolsep}{3.5pt}
\renewcommand{\arraystretch}{1.15}
\begin{tabular}{
|>{\raggedright\arraybackslash}p{0.26\linewidth}
|>{\raggedright\arraybackslash}p{0.38\linewidth}
|>{\raggedright\arraybackslash}p{0.28\linewidth}|}
\hline
Method family & Cached-memory interface & Role in our comparison \\
\hline
\textbf{Target-aware history attention} \newline
\emph{DIN, DIEN}
&
No fixed scorer-facing memory; the candidate interacts with user history during scoring.
&
Included as target-aware CTR accuracy baselines. \\
\hline

\textbf{Full-sequence backbones} \newline
\emph{MHA, SASRec, RetNet, Mamba4Rec, HSTU}
&
No fixed compact memory $\mathbf{H}_u$ by design; scoring depends on the retained sequence or implementation-specific state caching.
&
Included as sequence-model accuracy baselines. \\
\hline

\textbf{Two-tower matching} \newline
\emph{TTSN}
&
User embeddings can be precomputed, but the scorer is not the same target-conditioned compact-memory scorer.
&
Included as a matching-style accuracy baseline. \\
\hline

\textbf{Cached user-memory methods} \newline
\emph{VISTA, MARS}
&
Yes. Both expose compact scorer-facing memory $\mathbf{H}_u$ before candidate scoring. VISTA directly caches seed memories, while MARS constructs $\mathbf{H}_u$ from half-life-anchored temporal tracks and may cache $\mathcal{S}_u$ for incremental updates.
&
Serving-interface-matched family; VISTA is the primary apples-to-apples baseline. \\
\hline
\end{tabular}
\end{table}

\section{Public Dataset Protocol}
\label{app:public_protocol}

\paragraph{Dataset descriptions.}
Amazon-Electronics is a product recommendation dataset derived from Amazon reviews and item metadata, restricted to the electronics domain. Its histories are relatively short and serve as a short-history control.
KuaiRand-1K is a video recommendation dataset collected from Kuaishou logs with randomized exposure, providing longer user histories for CTR prediction.
KuaiRec is another recommendation dataset with longer histories, making it the primary public long-history benchmark in our experiments.

For all public datasets, user interactions are sorted chronologically.
Each CTR example uses only interactions before the target event as the user history, so future interactions are never exposed to the model. All methods use the same processed examples, feature fields, train/validation/test splits, and evaluation scripts.
The validation split is used for early stopping and model selection, and final public results are reported on the held-out test split over three random seeds. 

\paragraph{Compute resources.}
Public experiments were run on NVIDIA A100 80GB GPUs. Each public-dataset
training run completes in approximately 1--5 hours on a single A100. The
total compute for all main results, ablations, and analyses is approximately
800 GPU-hours.

\paragraph{Significance testing.}
We test significance between MARS and VISTA because VISTA is the primary serving-interface-matched cached-memory baseline.
For each public dataset, both models are evaluated on the same held-out test examples.
We use paired user-level bootstrap tests, where each bootstrap sample draws users with replacement, includes all test examples from the sampled users, and recomputes the AUC and NE differences between MARS and VISTA.

\paragraph{Dataset licenses and terms.}
KuaiRec and KuaiRand are released by their official repositories under the
CC-BY-SA-4.0 license. Amazon-Electronics follows the public
Amazon review data source and RecZoo preprocessing; the source page requests
research-purpose use and citation, while noting that it does not provide a
separate license for the underlying Amazon data. We use all public datasets for
non-commercial academic research, cite the original sources, and do not
redistribute any dataset.

\newpage

\section{Limitations}
\label{sec:limitations}

\paragraph{Scope.}
MARS targets long-history CTR-style recommendation in which a candidate-independent user memory is cached and reused across a large candidate pool.
Its gains are limited for short histories (Amazon-Electronics and the short-history bin in Figure~\ref{fig:analysis_all}(a)), and we make no claim about cold-start users.
All evaluations are offline; online deployment and other recommendation objectives are left for future work.

\paragraph{Structural hyperparameters.}
MARS introduces the number of temporal tracks, routing budget, memory depth, and half-life range. Performance is non-monotonic in the number of tracks, routing budget, and half-life range, and saturates with memory depth and seed capacity: too few tracks or too small an $H_{\max}$ limit temporal coverage, whereas too many tracks, near-dense readout, or a very large $H_{\max}$ reduce selectivity or mix in stale evidence (Section~\ref{subsec:ablation} and Appendix~\ref{app:hyper_sensitivity}).
Performance is stable around the default setting, but domains with very different activity rates may require re-tuning the half-life range.

\paragraph{Statistical evidence.}
Public results average three training runs, and our paired bootstrap tests quantify evaluation-set uncertainty rather than training-seed variability.
The industrial study (Appendix~\ref{app:industrial_validation}), the compressed state-bank variants, and several diagnostics are single-run, and the industrial results are reported only as relative NE changes on proprietary data.
The recoverability analyses use linear probes on a fixed cohort of long-history users and explain, rather than replace, recommendation metrics.

\paragraph{Efficiency and robustness evidence.}
Latency was measured for warm-cache serving on a single A100 platform and excludes one-time initialization from the raw history.
Event-level incremental updates require an optional per-user state, which increases storage unless it is compressed or restricted to active users.
The behavioral-shift analysis covers within-user shifts located by a model-independent drift score and does not establish robustness to arbitrary distribution shifts.

\paragraph{Broader risks.}
Like other long-history recommender models, MARS relies on user interaction histories~\citep{yin2024poisoning}, so production deployment should follow privacy regulations and monitor feedback loops, stale preferences, and behavioral overfitting.

\section*{NeurIPS Paper Checklist}

\begin{enumerate}

\item {\bf Claims}
\item[] Question: Do the main claims made in the abstract and introduction accurately reflect the paper's contributions and scope?
\item[] Answer: \answerYes{}
\item[] Justification: The abstract and Section~\ref{sec:intro} state the paper's scope: diagnosing temporal bottlenecks in cached long-history recommendation and proposing MARS as a multi-resolution user memory. The method is described in Section~\ref{sec:method}, and the empirical claims are supported in Section~\ref{sec:exp}.

\item {\bf Limitations}
\item[] Question: Does the paper discuss the limitations of the work performed by the authors?
\item[] Answer: \answerYes{}
\item[] Justification: Appendix~\ref{sec:limitations} discusses the scope (offline, long-history CTR prediction with cached user memory; limited gains for short histories; no cold-start claim), sensitivity to structural hyperparameters, the statistical scope of the evidence (bootstrap versus seed variability; single-run diagnostics; linear probes), the scope of the efficiency and behavioral-shift analyses, and deployment risks such as privacy and feedback loops.

\item {\bf Theory assumptions and proofs}
\item[] Question: For each theoretical result, does the paper provide the full set of assumptions and a complete proof?
\item[] Answer: \answerNA{}
\item[] Justification: The paper does not present formal theorems. The mathematical material consists of the model definition in Section~\ref{sec:method} and the parallel scan derivation in Appendix~\ref{app:parallel_scan}.

\item {\bf Experimental result reproducibility}
\item[] Question: Does the paper fully disclose all the information needed to reproduce the main experimental results to the extent that it affects the main claims and conclusions?
\item[] Answer: \answerYes{}
\item[] Justification: Section~\ref{subsec:setup} describes datasets, baselines, metrics, and evaluation protocol. Section~\ref{sec:method} specifies the architecture, Appendix~\ref{app:public_protocol} provides public-data protocol details, Table~\ref{tab:hyperparams} reports hyperparameter settings, and Appendices~\ref{app:realized_halflife}--\ref{app:latency} describe the protocols of the additional analyses.

\item {\bf Open access to data and code}
\item[] Question: Does the paper provide open access to the data and code, with sufficient instructions to faithfully reproduce the main experimental results?
\item[] Answer: \answerNo{}
\item[] Justification: The public datasets used in the paper are cited and linked in Section~\ref{subsec:setup}. The industrial dataset is proprietary and cannot be released; its results are supplementary, and the main claims rely on the public datasets. The code is not yet publicly available; the paper provides architectural, protocol, and hyperparameter details to support reproducibility, and we will release the implementation, preprocessing scripts, configurations, and evaluation code.

\item {\bf Experimental setting/details}
\item[] Question: Does the paper specify all the training and test details necessary to understand the results?
\item[] Answer: \answerYes{}
\item[] Justification: Section~\ref{subsec:setup} describes the public datasets, history budgets, baselines, metrics, and random-seed reporting. Appendix~\ref{app:public_protocol} details chronological construction and splits, and Appendix~\ref{app:hyper_sensitivity} lists the main hyperparameters.

\item {\bf Experiment statistical significance}
\item[] Question: Does the paper report error bars suitably and correctly defined or other appropriate information about statistical significance?
\item[] Answer: \answerYes{}
\item[] Justification: Table~\ref{tab:overall} reports mean and standard deviation over three runs for public datasets and reports paired user-level bootstrap significance between MARS and the interface-matched VISTA baseline. Component ablations, the behavioral-shift analysis, and the recoverability comparison report paired user-level bootstrap confidence intervals (Appendices~\ref{app:shift} and~\ref{app:component_details}), and single-run results, including the industrial study, are marked as such (e.g., Appendices~\ref{app:industrial_validation} and~\ref{app:compressed_state_cache}).
\item {\bf Experiments compute resources}
\item[] Question: For each experiment, does the paper provide sufficient information on the computer resources needed to reproduce the experiments?
\item[] Answer: \answerYes{}
\item[] Justification: Appendix~\ref{app:public_protocol} reports the GPU type,
approximate per-run training time, and total GPU-hours for the public
experiments.

\item {\bf Code of ethics}
\item[] Question: Does the research conducted in the paper conform, in every respect, with the NeurIPS Code of Ethics?
\item[] Answer: \answerYes{}
\item[] Justification: The paper studies offline recommendation modeling using public datasets and anonymized industrial aggregate results. It does not introduce human-subject experiments, deception, or high-risk data collection.

\item {\bf Broader impacts}
\item[] Question: Does the paper discuss both potential positive societal impacts and negative societal impacts of the work performed?
\item[] Answer: \answerYes{}
\item[] Justification: Appendix~\ref{sec:limitations} notes that practical recommender deployment should monitor privacy and feedback-loop effects. The positive impact is more accurate and efficient long-history recommendation, while the same deployment risks as other recommender systems remain.

\item {\bf Safeguards}
\item[] Question: Does the paper describe safeguards that have been put in place for responsible release of data or models that have a high risk for misuse?
\item[] Answer: \answerNA{}
\item[] Justification: The paper does not release a high-risk generative model, scraped dataset, or dual-use foundation model. It proposes a recommendation architecture evaluated offline.

\item {\bf Licenses for existing assets}
\item[] Question: Are the creators or original owners of assets used in the paper properly credited, and are licenses and terms of use explicitly mentioned and respected?
\item[] Answer: \answerYes{}
\item[] Justification: Public datasets and baselines are cited where introduced.
Dataset licenses or source terms, and our non-redistribution practice, are
summarized in Appendix~\ref{app:public_protocol}.

\item {\bf New assets}
\item[] Question: Are new assets introduced in the paper well documented and is the documentation provided alongside the assets?
\item[] Answer: \answerNA{}
\item[] Justification: The submission does not introduce or release a new dataset, benchmark, or code asset.

\item {\bf Crowdsourcing and research with human subjects}
\item[] Question: For crowdsourcing experiments and research with human subjects, does the paper include the required participant instructions, screenshots, and compensation details?
\item[] Answer: \answerNA{}
\item[] Justification: The paper does not involve crowdsourcing or new research with human subjects.

\item {\bf Institutional review board approvals or equivalent for research with human subjects}
\item[] Question: Does the paper describe potential risks to study participants and whether IRB approvals or equivalent review were obtained?
\item[] Answer: \answerNA{}
\item[] Justification: The paper does not involve new research with human subjects.

\item {\bf Declaration of LLM usage}
\item[] Question: Does the paper describe the usage of LLMs if it is an important, original, or non-standard component of the core methods in this research?
\item[] Answer: \answerNA{}
\item[] Justification: LLMs are not used as part of the MARS architecture, experiments, or scientific methodology.

\end{enumerate}

\end{document}